%% file: main.tex
\documentclass[a4paper]{article}
\usepackage[bottom=2.5cm]{geometry}

\usepackage[inline]{enumitem}
\usepackage{acronym}
\usepackage[capitalise]{cleveref}
\usepackage[sort&compress,numbers]{natbib}
\usepackage{xcolor}
\usepackage{multirow}
\usepackage{booktabs}
\usepackage{tcolorbox}
\usepackage{authblk}
\usepackage{url}
\usepackage{amssymb}
\usepackage{fancyhdr}
\newcommand{\editorMark}{\textsuperscript{*}}

\newcommand{\authorrunning}{Bauer et al.}
\newcommand{\titlerunning}{Dagstuhl Perspectives Workshop 24352---Conversational Agents: A Framework for Evaluation (CAFE): Manifesto}

\definecolor{block-gray}{gray}{0.93}
\newtcolorbox{citationblock}{colback=block-gray,boxrule=0pt,boxsep=0pt}

\fancypagestyle{plain}{
  \fancyhf{}
  \fancyfoot[C]{\thepage}
}

\DeclareGraphicsExtensions{.pdf,.ai,.jpg,.png}
\graphicspath{{./figure/}}
\setkeys{Gin}{pagebox=artbox}

\begin{document}

\title{Dagstuhl Perspectives Workshop 24352---Conversational Agents: A Framework for Evaluation (CAFE): Manifesto}

\author[1]{Christine Bauer\editorMark}
\author[2]{Li Chen\editorMark}
\author[3]{Nicola Ferro\editorMark}
\author[4]{Norbert Fuhr\editorMark}

\affil[1]{University of Salzburg, AT \\ \texttt{christine.bauer@plus.ac.at}}
\affil[2]{Hong Kong Baptist University, HK, China \\ \texttt{lichen@comp.hkbu.edu.hk}}
\affil[3]{University of Padua, IT \\ \texttt{nicola.ferro@unipd.it}}
\affil[4]{Universität Duisburg-Essen, DE \\ \texttt{norbert.fuhr@uni-due.de}}

\author[5]{Avishek Anand}
\affil[5]{TU Delft, NL \\ \texttt{avishek.anand@tudelft.nl}}

\author[6]{Timo Breuer}
\affil[6]{TH Köln, DE \\ \texttt{timo.breuer@th-koeln.de}}

\author[7]{Guglielmo Faggioli}
\affil[7]{University of Padua, IT \\ \texttt{guglielmo.faggioli@unipd.it}}

\author[8]{Ophir Frieder}
\affil[8]{Georgetown University, USA \\ \texttt{ophir@ir.cs.georgetown.edu}}

\author[9]{Hideo Joho}
\affil[9]{University of Tsukuba, JP \\ \texttt{hideo@slis.tsukuba.ac.jp}}

\author[10]{Jussi Karlgren}
\affil[10]{Silo AI, FI \\ \texttt{jussi.karlgren@silo.ai}}

\author[11]{Johannes Kiesel}
\affil[11]{Bauhaus-Universität Weimar, DE \\ \texttt{johannes.kiesel@uni-weimar.de}}

\author[12]{Bart P. Knijnenburg}
\affil[12]{Clemson University, USA \\ \texttt{bartk@clemson.edu}}

\author[13]{Aldo Lipani}
\affil[13]{University College London, UK \\ \texttt{aldo.lipani@ucl.ac.uk}}

\author[14]{Lien Michiels}
\affil[14]{imec-SMIT, Vrije Universiteit Brussel \& University of Antwerp, BE \\ \texttt{lien.michiels@uantwerpen.be}}

\author[15]{Andrea Papenmeier}
\affil[15]{University of Twente, NL \\ \texttt{a.papenmeier@utwente.nl}}

\author[16]{Maria Soledad Pera}
\affil[16]{TU Delft, NL \\ \texttt{m.s.pera@tudelft.nl}}

\author[17]{Mark Sanderson}
\affil[17]{RMIT University, AU \\ \texttt{mark.sanderson@rmit.edu.au}}

\author[18]{Scott Sanner}
\affil[18]{University of Toronto, CA \\ \texttt{ssanner@mie.utoronto.ca}}

\author[19]{Benno Stein}
\affil[19]{Bauhaus-Universität Weimar, DE \\ \texttt{benno.stein@uni-weimar.de}}

\author[20]{Johanne R. Trippas}
\affil[20]{RMIT University, AU \\ \texttt{j.trippas@rmit.edu.au}}

\author[21]{Karin Verspoor}
\affil[21]{RMIT University, AU \\ \texttt{karin.verspoor@rmit.edu.au}}

\author[22]{Martijn C. Willemsen}
\affil[22]{TU Eindhoven \& JADS, NL \\ \texttt{m.c.willemsen@tue.nl}}

\date{}

\maketitle

\begingroup
\renewcommand{\thefootnote}{\fnsymbol{footnote}}
\footnotetext[1]{Authors marked with * are editors.}
\endgroup

\thispagestyle{plain} 

\input{./used_acronyms}

\begin{citationblock}
\flushleft
\vspace{5pt}
This is the accepted version of the ``Manifesto from Dagstuhl Perspectives Workshop~24352---Conversational Agents: A Framework for Evaluation (CAFE)'' (before final copyediting). This Dagstuhl Perspectives Workshop~24352 was held {25.--30.~August, 2024}. For details, see \url{https://www.dagstuhl.de/24352}.

\vspace{5pt}
Cite as: 

Christine Bauer, Li Chen, Nicola Ferro, Norbert Fuhr, Avishek Anand, Timo Breuer, Guglielmo Faggioli, Ophir Frieder, Hideo Joho, Jussi Karlgren, Johannes Kiesel, Bart P. Knijnenburg, Aldo Lipani, Lien Michiels, Andrea Papenmeier, Maria Soledad Pera, Mark Sanderson, Scott Sanner, Benno Stein, Johanne R. Trippas, Karin Verspoor, and Martijn C. Willemsen. Conversational Agents: A Framework for Evaluation (CAFE) (Dagstuhl Perspectives Workshop 24352). In Dagstuhl Manifestos, Volume 11, Issue 1, pp. 19--67, Schloss Dagstuhl -- Leibniz-Zentrum für Informatik (2025). \url{https://doi.org/10.4230/DagMan.11.1.19}
\vspace{5pt}
\end{citationblock}
\vspace{20pt}

\input{section/abstract}

\input{section/executive-summary}

\input{section/introduction}

\input{section/coniac}

\input{section/cafe}

\input{section/goal}

\input{section/user}

\input{section/task}

\input{section/criteria}

\input{section/methodology}

\input{section/measure}

\input{section/research}

\input{section/conclusions}


\input{section/acknowledgements}

\bibliography{cited}

\end{document}

%% file: used_acronyms.tex
\acrodef{3G}[3G]{Third Generation Mobile System}
\acrodef{5S}[5S]{Streams, Structures, Spaces, Scenarios, Societies}
\acrodef{AA}[AA]{Active Agreements}
\acrodef{AAAI}[AAAI]{Association for the Advancement of Artificial Intelligence}
\acrodef{AAL}[AAL]{Annotation Abstraction Layer}
\acrodef{AAM}[AAM]{Automatic Annotation Manager}
\acrodef{AAP}[AAP]{Average Average Precision}
\acrodef{ACLIA}[ACLIA]{Advanced Cross-Lingual Information Access}
\acrodef{ACM}[ACM]{Association for Computing Machinery}
\acrodef{AD}[AD]{Active Disagreements}
\acrodef{ADSL}[ADSL]{Asymmetric Digital Subscriber Line}
\acrodef{ADUI}[ADUI]{ADministrator User Interface}
\acrodef{AGI}[AGI]{Artificial General Intelligence}
\acrodef{AI}[AI]{Artificial Intelligence}
\acrodef{AIP}[AIP]{Archival Information Package}
\acrodef{AJAX}[AJAX]{Asynchronous JavaScript Technology and \acs{XML}}
\acrodef{ALS}[ALS]{Amyotrophic Lateral Sclerosis}
\acrodef{ALSFRS-R}[ALSFRS-R]{ALS Functional Rating Scale Revisited}
\acrodef{ALU}[ALU]{Aritmetic-Logic Unit}
\acrodef{AMP}[AMP]{Axiomatic Model of Preferences}
\acrodef{AMUSID}[AMUSID]{Adaptive MUSeological IDentity-service}
\acrodef{ANOVA}[ANOVA]{ANalysis Of VAriance}
\acrodef{ANSI}[ANSI]{American National Standards Institute}
\acrodef{AP}[AP]{Average Precision}
\acrodef{APC}[APC]{AP Correlation}
\acrodef{API}[API]{Application Program Interface}
\acrodef{AR}[AR]{Address Register}
\acrodef{AS}[AS]{Annotation Service}
\acrodef{ASAP}[ASAP]{Adaptable Software Architecture Performance}
\acrodef{ASI}[ASI]{Annotation Service Integrator}
\acrodef{ASL}[ASL]{Achieved Significance Level}
\acrodef{ASM}[ASM]{Annotation Storing Manager}
\acrodef{ASR}[ASR]{Automatic Speech Recognition}
\acrodef{ASUI}[ASUI]{ASsessor User Interface}
\acrodef{ATIM}[ATIM]{Annotation Textual Indexing Manager}
\acrodef{AUC}[AUC]{Area Under the ROC Curve}
\acrodef{AUI}[AUI]{Administrative User Interface}
\acrodef{AWARE}[AWARE]{Assessor-driven Weighted Averages for Retrieval Evaluation}
\acrodef{BANKS-I}[BANKS-I]{Browsing ANd Keyword Searching I}
\acrodef{BANKS-II}[BANKS-II]{Browsing ANd Keyword Searching II}
\acrodef{BBH}[BBH]{BIG-Bench Hard}
\acrodef{BH}[BH]{Benjamini-Hochberg}
\acrodef{bpref}[bpref]{Binary Preference}
\acrodef{BNF}[BNF]{Backus and Naur Form}
\acrodef{BPM}[BPM]{Bejeweled Player Model}
\acrodef{BRICKS}[BRICKS]{Building Resources for Integrated Cultural Knowledge Services}
\acrodef{CAFE}[CAFE]{Conversational Agents Framework for Evaluation}
\acrodef{CAN}[CAN]{Content Addressable Netword}
\acrodef{CAS}[CAS]{Content-And-Structure}
\acrodef{CBSD}[CBSD]{Component-Based Software Developlement}
\acrodef{CBSE}[CBSE]{Component-Based Software Engineering}
\acrodef{CB-SPE}[CB-SPE]{Component-Based \acs{SPE}}
\acrodef{CD}[CD]{Collaboration Diagram}
\acrodef{CD}[CD]{Compact Disk}
\acrodef{CDF}[CDF]{Cumulative Density Function}
\acrodef{CENL}[CENL]{Conference of European National Librarians}
\acrodef{ChatGPT}[ChatGPT]{Chat Generative Pre-trained Transformer}
\acrodef{CONIAC}[CONIAC]{CONversational Information ACcess} 
\acrodef{CIDOC CRM}[CIDOC CRM]{CIDOC Conceptual Reference Model}
\acrodef{CIR}[CIR]{Current Instruction Register}
\acrodef{CIRCO}[CIRCO]{Coordinated Information Retrieval Components Orchestration}
\acrodef{CIS}[CIS]{Conversational Information Seeking}
\acrodef{CG}[CG]{Cumulated Gain}
\acrodef{CL}[CL]{Curriculum Learning}
\acrodef{CL-ESA}[CL-ESA]{Cross-Lingual Explicit Semantic Analysis}
\acrodef{CLAIRE}[CLAIRE]{Combinatorial visuaL Analytics system for Information Retrieval Evaluation}
\acrodef{CLEF1}[CLEF]{Cross-Language Evaluation Forum}
\acrodef{CLEF}[CLEF]{Conference and Labs of the Evaluation Forum}
\acrodef{CLIR}[CLIR]{Cross Language Information Retrieval}
\acrodef{CM}[CM]{Continuation Methods}
\acrodef{CMS}[CMS]{Content Management System}
\acrodef{CMT}[CMT]{Campaign Management Tool}
\acrodef{CNR}[CNR]{Italian National Council of Research}
\acrodef{CO}[CO]{Content-Only}
\acrodef{COD}[COD]{Code On Demand}
\acrodef{CODATA}[CODATA]{Committee on Data for Science and Technology}
\acrodef{COLLATE}[COLLATE]{Collaboratory for Annotation Indexing and Retrieval of Digitized Historical Archive Material}
\acrodef{CP}[CP]{Characteristic Pattern}
\acrodef{CPE}[CPE]{Control Processor Element}
\acrodef{CPU}[CPU]{Central Processing Unit}
\acrodef{CQL}[CQL]{Contextual Query Language}
\acrodef{CRP}[CRP]{Cumulated Relative Position}
\acrodef{CRUD}[CRUD]{Create--Read--Update--Delete}
\acrodef{CS}[CS]{Characteristic Structure}
\acrodef{CSM}[CSM]{Campaign Storing Manager}
\acrodef{CSS}[CSS]{Cascading Style Sheets}
\acrodef{CTR}[CTR]{Click-Through Rate}
\acrodef{CU}[CU]{Control Unit}
\acrodef{CUI}[CUI]{Client User Interface}
\acrodef{CV}[CV]{Cross-Validation}
\acrodef{DAFFODIL}[DAFFODIL]{Distributed Agents for User-Friendly Access of Digital Libraries}
\acrodef{DAO}[DAO]{Data Access Object}
\acrodef{DARE}[DARE]{Drawing Adequate REpresentations}
\acrodef{DARPA}[DARPA]{Defense Advanced Research Projects Agency}
\acrodef{DAS}[DAS]{Distributed Annotation System}
\acrodef{DB}[DB]{DataBase}
\acrodef{DBMS}[DBMS]{DataBase Management System}
\acrodef{DC}[DC]{Dublin Core}
\acrodef{DCG}[DCG]{Discounted Cumulated Gain}
\acrodef{DCMI}[DCMI]{Dublin Core Metadata Initiative}
\acrodef{DCV}[DCV]{Document Cut--off Value}
\acrodef{DD}[DD]{Deployment Diagram}
\acrodef{DDC}[DDC]{Dewey Decimal Classification}
\acrodef{DDS}[DDS]{Direct Data Structure}
\acrodef{DF}[DF]{Degrees of Freedom}
\acrodef{DFI}[DFI]{Divergence From Independence}
\acrodef{DFR}[DFR]{Divergence From Randomness}
\acrodef{DHT}[DHT]{Distributed Hash Table}
\acrodef{DI}[DI]{Digital Image}
\acrodef{DIKW}[DIKW]{Data, Information, Knowledge, Wisdom}
\acrodef{DIL}[DIL]{\acs{DIRECT} Integration Layer}
\acrodef{DiLAS}[DiLAS]{Digital Library Annotation Service}
\acrodef{DIRECT}[DIRECT]{Distributed Information Retrieval Evaluation Campaign Tool}
\acrodef{DKMS}[DKMS]{Data and Knowledge Management System}
\acrodef{DL}[DL]{Digital Library}
\acrodefplural{DL}[DL]{Digital Libraries}
\acrodef{DLMS}[DLMS]{Digital Library Management System}
\acrodef{DLOG}[DL]{Description Logics}
\acrodef{DLS}[DLS]{Digital Library System}
\acrodef{DLSS}[DLSS]{Digital Library Service System}
\acrodef{DM}[DM]{Data Mining}
\acrodef{DO}[DO]{Digital Object}
\acrodef{DOI}[DOI]{Digital Object Identifier}
\acrodef{DOM}[DOM]{Document Object Model}
\acrodef{DoMDL}[DoMDL]{Document Model for Digital Libraries}
\acrodef{DP}[DP]{Discriminative Power}
\acrodef{DPBF}[DPBF]{Dynamic Programming Best-First}
\acrodef{DR}[DR]{Data Register}
\acrodef{DRIVER}[DRIVER]{Digital Repository Infrastructure Vision for European Research}
\acrodef{DTD}[DTD]{Document Type Definition}
\acrodef{DVD}[DVD]{Digital Versatile Disk}
\acrodef{EAC-CPF}[EAC-CPF]{Encoded Archival Context for Corporate Bodies, Persons, and Families}
\acrodef{EAD}[EAD]{Encoded Archival Description}
\acrodef{EAN}[EAN]{International Article Number}
\acrodef{EBU}[EBU]{Expected Browsing Utility}
\acrodef{ECD}[ECD]{Enhanced Contenty Delivery}
\acrodef{ECDL}[ECDL]{European Conference on Research and Advanced Technology for Digital Libraries}
\acrodef{EDM}[EDM]{Europeana Data Model}
\acrodef{EG}[EG]{Execution Graph}
\acrodef{ELDA}[ELDA]{Evaluation and Language resources Distribution Agency}
\acrodef{ELRA}[ELRA]{European Language Resources Association}
\acrodef{EM}[EM]{Expectation Maximization}
\acrodef{EMMA}[EMMA]{Extensible MultiModal Annotation}
\acrodef{EPROM}[EPROM]{Erasable Programmable \acs{ROM}}
\acrodef{EQNM}[EQNM]{Extended Queueing Network Model}
\acrodef{ER}[ER]{Entity--Relationship}
\acrodef{ERR}[ERR]{Expected Reciprocal Rank}
\acrodef{ERS}[ERS]{Empirical Relational System}
\acrodef{ESA}[ESA]{Explicit Semantic Analysis}
\acrodef{ESL}[ESL]{Expected Search Length}
\acrodef{ETL}[ETL]{Extract-Transform-Load}
\acrodef{FAST}[FAST]{Flexible Annotation Service Tool}
\acrodef{FDR}[FDR]{False Discovery Rate}
\acrodef{FIFO}[FIFO]{First-In / First-Out}
\acrodef{FIRE}[FIRE]{Forum for Information Retrieval Evaluation}
\acrodef{FN}[FN]{False Negative}
\acrodef{FNR}[FNR]{False Negative Rate}
\acrodef{FOAF}[FOAF]{Friend of a Friend}
\acrodef{FORESEE}[FORESEE]{FOod REcommentation sErvER}
\acrodef{FP}[FP]{False Positive}
\acrodef{FPC}[FPC]{Finite Population Correction}
\acrodef{FPR}[FPR]{False Positive Rate}
\acrodef{FVC}[FVC]{Forced Vital Capacity}
\acrodef{FWER}[FWER]{Family-wise Error Rate}
\acrodef{GIF}[GIF]{Graphics Interchange Format}
\acrodef{GIR}[GIR]{Geografic Information Retrieval}
\acrodef{GAP}[GAP]{Graded Average Precision}
\acrodef{GLM}[GLM]{General Linear Model}
\acrodef{GLMM}[GLMM]{General Linear Mixed Model}
\acrodef{GMAP}[GMAP]{Geometric Mean Average Precision}
\acrodef{GoP}[GoP]{Grid of Points}
\acrodef{GPRS}[GPRS]{General Packet Radio Service}
\acrodef{GPT}[GPT]{Generative Pre-trained Transformer}
\acrodef{gP}[gP]{Generalized Precision}
\acrodef{gR}[gR]{Generalized Recall}
\acrodef{gRBP}[gRBP]{Graded Rank-Biased Precision}
\acrodef{GT}[GT]{Generalizability Theory}
\acrodef{GTIN}[GTIN]{Global Trade Item Number}
\acrodef{GUI}[GUI]{Graphical User Interface}
\acrodef{GW}[GW]{Gateway}
\acrodef{HCI}[HCI]{Human Computer Interaction}
\acrodef{HDS}[HDS]{Hybrid Data Structure}
\acrodef{HIR}[HIR]{Hypertext Information Retrieval}
\acrodef{HIT}[HIT]{Human Intelligent Task}
\acrodef{HITS}[HITS]{Hyperlink-Induced Topic Search}
\acrodef{HMM}[HMM]{Hidden Markov Model}
\acrodef{HTML}[HTML]{HyperText Markup Language}
\acrodef{HTTP}[HTTP]{HyperText Transfer Protocol}
\acrodef{HSD}[HSD]{Honestly Significant Difference}
\acrodef{ICA}[ICA]{International Council on Archives}
\acrodef{ICSU}[ICSU]{International Council for Science}
\acrodef{IDF}[IDF]{Inverse Document Frequency}
\acrodef{iDPP}[iDPP@CLEF]{Intelligent Disease Progression Prediction at CLEF}
\acrodef{IDS}[IDS]{Inverse Data Structure}
\acrodef{IEEE}[IEEE]{Institute of Electrical and Electronics Engineers}
\acrodef{IEI}[IEI]{Istituto della Enciclopedia Italiana fondata da Giovanni Treccani}
\acrodef{IETF}[IETF]{Internet Engineering Task Force}
\acrodef{IIR}[IIR]{Interactive Information Retrieval}
\acrodef{IMS}[IMS]{Information Management System}
\acrodef{IMSPD}[IMS]{Information Management Systems Research Group}
\acrodef{indAP}[indAP]{Induced Average Precision}
\acrodef{infAP}[infAP]{Inferred Average Precision}
\acrodef{INEX}[INEX]{INitiative for the Evaluation of \acs{XML} Retrieval}
\acrodef{INS-M}[INS-M]{Inverse Set Data Model}
\acrodef{INTR}[INTR]{Interrupt Register}
\acrodef{IP}[IP]{Internet Protocol}
\acrodef{IPSA}[IPSA]{Imaginum Patavinae Scientiae Archivum}
\acrodef{IR}[IR]{Information Retrieval}
\acrodef{IRON}[IRON]{Information Retrieval ON}
\acrodef{IRON2}[IRON$^2$]{Information Retrieval On aNNotations}
\acrodef{IRON-SAT}[IRON-SAT]{\acs{IRON}---Statistical Analysis Tool}
\acrodef{IS}[IS]{Information Seeking}
\acrodef{IRS}[IRS]{Information Retrieval System}
\acrodef{ISAD(G)}[ISAD(G)]{International Standard for Archival Description (General)}
\acrodef{ISBN}[ISBN]{International Standard Book Number}
\acrodef{ISIS}[ISIS]{Interactive SImilarity Search}
\acrodef{ISJ}[ISJ]{Interactive Searching and Judging}
\acrodef{ISO}[ISO]{International Organization for Standardization}
\acrodef{ITU}[ITU]{International Telecommunication Union }
\acrodef{ITU-T}[ITU-T]{Telecommunication Standardization Sector of \acs{ITU}}
\acrodef{IV}[IV]{Information Visualization}
\acrodef{JAN}[JAN]{Japanese Article Number}
\acrodef{JDBC}[JDBC]{Java DataBase Connectivity}
\acrodef{JMB}[JMB]{Java--Matlab Bridge}
\acrodef{JPEG}[JPEG]{Joint Photographic Experts Group}
\acrodef{JSON}[JSON]{JavaScript Object Notation}
\acrodef{JSP}[JSP]{Java Server Pages}
\acrodef{JTE}[JTE]{Java-Treceval Engine}
\acrodef{KDE}[KDE]{Kernel Density Estimation}
\acrodef{KLD}[KLD]{Kullback-Leibler Divergence}
\acrodef{KLAPER}[KLAPER]{Kernel LAnguage for PErformance and Reliability analysis}
\acrodef{LAM}[LAM]{Libraries, Archives, and Museums}
\acrodef{LAM2}[LAM]{Logistic Average Misclassification}
\acrodef{LAN}[LAN]{Local Area Network}
\acrodef{LD}[LD]{Linked Data}
\acrodef{LEAF}[LEAF]{Linking and Exploring Authority Files}
\acrodef{LIDO}[LIDO]{Lightweight Information Describing Objects}
\acrodef{LIFO}[LIFO]{Last-In / First-Out}
\acrodef{LLaMA}[LLaMA]{Large Language Model Meta AI}
\acrodef{LM}[LM]{Language Model}
\acrodef{LLM}[LLM]{Large Language Model}
\acrodef{LLMs}[LLMs]{Large Language Models}
\acrodef{LMT}[LMT]{Log Management Tool}
\acrodef{LOD}[LOD]{Linked Open Data}
\acrodef{LODE}[LODE]{Linking Open Descriptions of Events}
\acrodef{LpO}[LpO]{Leave-$p$-Out}
\acrodef{LRM}[LRM]{Local Relational Model}
\acrodef{LRU}[LRU]{Last Recently Used}
\acrodef{LS}[LS]{Lexical Signature}
\acrodef{LSM}[LSM]{Log Storing Manager}
\acrodef{LtR}[LtR]{Learning to Rank}
\acrodef{LUG}[LUG]{Lexical Unit Generator}
\acrodef{MA}[MA]{Mobile Agent}
\acrodef{MA}[MA]{Moving Average}
\acrodef{MACS}[MACS]{Multilingual ACcess to Subjects}
\acrodef{MADCOW}[MADCOW]{Multimedia Annotation of Digital Content Over the Web}
\acrodef{MAD}[MAD]{Mean Assessed Documents}
\acrodef{MADP}[MADP]{Mean Assessed Documents Precision}
\acrodef{MADS}[MADS]{Metadata Authority Description Standard}
\acrodef{MAP}[MAP]{Mean Average Precision}
\acrodef{MARC}[MARC]{Machine Readable Cataloging}
\acrodef{MATTERS}[MATTERS]{MATlab Toolkit for Evaluation of information Retrieval Systems}
\acrodef{MDA}[MDA]{Model Driven Architecture}
\acrodef{MDD}[MDD]{Model-Driven Development}
\acrodef{METS}[METS]{Metadata Encoding and Transmission Standard}
\acrodef{MIDI}[MIDI]{Musical Instrument Digital Interface}
\acrodef{MIME}[MIME]{Multipurpose Internet Mail Extensions}
\acrodef{ML}[ML]{Machine Learning}
\acrodef{MLE}[MLE]{Maximum Likelihood Estimation}
\acrodef{MLIA}[MLIA]{MultiLingual Information Access}
\acrodef{MM}[MM]{Machinery Model}
\acrodef{MMLU}[MMLU]{Massive Multitask Language Understanding}
\acrodef{MMU}[MMU]{Memory Management Unit}
\acrodef{MODS}[MODS]{Metadata Object Description Schema}
\acrodef{MOF}[MOF]{Meta-Object Facility}
\acrodef{MP}[MP]{Markov Precision}
\acrodef{MPEG}[MPEG]{Motion Picture Experts Group}
\acrodef{MRD}[MRD]{Machine Readable Dictionary}
\acrodef{MRF}[MRF]{Markov Random Field}
\acrodef{MRR}[MRR]{Mean Reciprocal Rank}
\acrodef{MS}[MS]{Mean Squares}
\acrodef{MS2}[MS]{Multiple Sclerosis}
\acrodef{MSAC}[MSAC]{Multilingual Subject Access to Catalogues}
\acrodef{MSE}[MSE]{Mean Square Error}
\acrodef{MT}[MT]{Machine Translation}
\acrodef{MV}[MV]{Majority Vote}
\acrodef{MVC}[MVC]{Model-View-Controller}
\acrodef{NACSIS}[NACSIS]{NAtional Center for Science Information Systems}
\acrodef{NAP}[NAP]{Network processors Applications Profile}
\acrodef{NASEM}[NASEM]{National Academies of Sciences, Engineering, and Medicine}
\acrodef{NCP}[NCP]{Normalized Cumulative Precision}
\acrodef{nCG}[nCG]{Normalized Cumulated Gain}
\acrodef{nCRP}[nCRP]{Normalized Cumulated Relative Position}
\acrodef{nDCG}[nDCG]{Normalized Discounted Cumulated Gain}
\acrodef{nMCG}[nMCG]{Normalized Markov Cumulated Gain}
\acrodef{NESTOR}[NESTOR]{NEsted SeTs for Object hieRarchies}
\acrodef{NEXI}[NEXI]{Narrowed Extended XPath I}
\acrodef{NII}[NII]{National Institute of Informatics}
\acrodef{NISO}[NISO]{National Information Standards Organization}
\acrodef{NIST}[NIST]{National Institute of Standards and Technology}
\acrodef{NIV}[NIV]{Non-Invasive Ventilation}
\acrodef{NLP}[NLP]{Natural Language Processing}
\acrodef{NN}[NN]{Neural Network}
\acrodef{NP}[NP]{Network Processor}
\acrodef{NR}[NR]{Normalized Recall}
\acrodef{NRS}[NRS]{Numerical Relational System}
\acrodef{NS-M}[NS-M]{Nested Set Model}
\acrodef{NTCIR}[NTCIR]{NII Testbeds and Community for Information access Research}
\acrodef{OAI}[OAI]{Open Archives Initiative}
\acrodef{OAI-ORE}[OAI-ORE]{Open Archives Initiative Object Reuse and Exchange}
\acrodef{OAI-PMH}[OAI-PMH]{Open Archives Initiative Protocol for Metadata Harvesting}
\acrodef{OAIS}[OAIS]{Open Archival Information System}
\acrodef{OC}[OC]{Operation Code}
\acrodef{OCLC}[OCLC]{Online Computer Library Center}
\acrodef{OMG}[OMG]{Object Management Group}
\acrodef{OO}[OO]{Object Oriented}
\acrodef{OODB}[OODB]{Object-Oriented \acs{DB}}
\acrodef{OODBMS}[OODBMS]{Object-Oriented \acs{DBMS}}
\acrodef{OPAC}[OPAC]{Online Public Access Catalog}
\acrodef{OQL}[OQL]{Object Query Language}
\acrodef{ORP}[ORP]{Open Relevance Project}
\acrodef{OSIRIS}[OSIRIS]{Open Service Infrastructure for Reliable and Integrated process Support}
\acrodef{P}[P]{Precision}
\acrodef{P2P}[P2P]{Peer-To-Peer}
\acrodef{PA}[PA]{Passive Agreements}
\acrodef{PAMT}[PAMT]{Pool-Assessment Management Tool}
\acrodef{PASM}[PASM]{Pool-Assessment Storing Manager}
\acrodef{PC}[PC]{Program Counter}
\acrodef{PCP}[PCP]{Pre-Commercial Procurement}
\acrodef{PCR}[PCR]{Peripherical Command Register}
\acrodef{PD}[PD]{Passive Disagreements}
\acrodef{PDA}[PDA]{Personal Digital Assistant}
\acrodef{PDF}[PDF]{Probability Density Function}
\acrodef{PDR}[PDR]{Peripherical Data Register}
\acrodef{PEG}[PEG]{Percutaneous Endoscopic Gastrostomy}
\acrodef{PIR}[PIR]{Personalized Information Retrieval}
\acrodef{POI}[POI]{\acs{PURL}-based Object Identifier}
\acrodef{POMDP}[POMDP]{Partially Observable Markov Decision Process}
\acrodef{PoS}[PoS]{Part of Speech}
\acrodef{PAA}[PAA]{Proportion of Active Agreements}
\acrodef{PPA}[PPA]{Proportion of Passive Agreements}
\acrodef{PPE}[PPE]{Programmable Processing Engine}
\acrodef{PREFORMA}[PREFORMA]{PREservation FORMAts for culture information/e-archives}
\acrodef{PRIMAD}[PRIMAD]{Platform, Research goal, Implementation, Method, Actor, and Data}
\acrodef{PRIMAmob-UML}[PRIMAmob-UML]{mobile \acs{PRIMA-UML}}
\acrodef{PRIMA-UML}[PRIMA-UML]{PeRformance IncreMental vAlidation in \acs{UML}}
\acrodef{PROM}[PROM]{Programmable \acs{ROM}}
\acrodef{PROMISE}[PROMISE]{Participative Research labOratory for Multimedia and Multilingual Information Systems Evaluation}
\acrodef{pSQL}[pSQL]{propagate \acs{SQL}}
\acrodef{PUI}[PUI]{Participant User Interface}
\acrodef{PURL}[PURL]{Persistent \acs{URL}}
\acrodef{QA}[QA]{Question Answering}
\acrodef{QC}[QC]{Quantum Computing}
\acrodef{QCA}[QA]{Quantum Annealing}
\acrodef{QE}[QE]{Query Expansion}
\acrodef{QoS-UML}[QoS-UML]{\acs{UML} Profile for QoS and Fault Tolerance}
\acrodef{QPA}[QPA]{Query Performance Analyzer}
\acrodef{QPP}[QPP]{Query Performance Prediction}
\acrodef{R}[R]{Recall}
\acrodef{RAG}[RAG]{Retrieval-Augmented Generation}
\acrodef{RAM}[RAM]{Random Access Memory}
\acrodef{RAMM}[RAM]{Random Access Machine}
\acrodef{RBO}[RBO]{Rank-Biased Overlap}
\acrodef{RBP}[RBP]{Rank-Biased Precision}
\acrodef{RBTO}[RBTO]{Rank-Based Total Order}
\acrodef{RDBMS}[RDBMS]{Relational \acs{DBMS}}
\acrodef{RDF}[RDF]{Resource Description Framework}
\acrodef{REST}[REST]{REpresentational State Transfer}
\acrodef{REV}[REV]{Remote Evaluation}
\acrodef{RF}[RF]{Relevance Feedback}
\acrodef{RFC}[RFC]{Request for Comments}
\acrodef{RIA}[RIA]{Reliable Information Access}
\acrodef{RMSE}[RMSE]{Root Mean Square Error}
\acrodef{RMT}[RMT]{Run Management Tool}
\acrodef{ROM}[ROM]{Read Only Memory}
\acrodef{ROMIP}[ROMIP]{Russian Information Retrieval Evaluation Seminar}
\acrodef{RoMP}[RoMP]{Rankings of Measure Pairs}
\acrodef{RoS}[RoS]{Rankings of Systems}
\acrodef{RP}[RP]{Relative Position}
\acrodef{RR}[RR]{Reciprocal Rank}
\acrodef{RS}[RS]{Recommender Systems}
\acrodef{RSM}[RSM]{Run Storing Manager}
\acrodef{RST}[RST]{Rhetorical Structure Theory}
\acrodef{RSV}[RSV]{Retrieval Status Value}
\acrodef{RT-UML}[RT-UML]{\acs{UML} Profile for Schedulability, Performance and Time}
\acrodef{SA}[SA]{Software Architecture}
\acrodef{SAL}[SAL]{Storing Abstraction Layer}
\acrodef{SAMT}[SAMT]{Statistical Analysis Management Tool}
\acrodef{SAN}[SAN]{Sistema Archivistico Nazionale}
\acrodef{SASM}[SASM]{Statistical Analysis Storing Manager}
\acrodef{SBTO}[SBTO]{Set-Based Total Order}
\acrodef{SD}[SD]{Sequence Diagram}
\acrodef{SE}[SE]{Search Engine}
\acrodef{SEBD}[SEBD]{Convegno Nazionale su Sistemi Evoluti per Basi di Dati}
\acrodef{SEM}[SEM]{Standard Error of the Mean}
\acrodef{SERP}[SERP]{Search Engine Result Page}
\acrodef{SFT}[SFT]{Satisfaction--Frustration--Total}
\acrodef{SIL}[SIL]{Service Integration Layer}
\acrodef{SIP}[SIP]{Submission Information Package}
\acrodef{SKOS}[SKOS]{Simple Knowledge Organization System}
\acrodef{SM}[SM]{Software Model}
\acrodef{SME}[SME]{Statistics--Metrics-Experiments}
\acrodef{SMART}[SMART]{System for the Mechanical Analysis and Retrieval of Text}
\acrodef{SoA}[SoA]{Service-oriented Architectures}
\acrodef{SOA}[SOA]{Strength of Association}
\acrodef{SOAP}[SOAP]{Simple Object Access Protocol}
\acrodef{SOM}[SOM]{Self-Organizing Map}
\acrodef{SPARQL}[SPARQL]{Simple Protocol and RDF Query Language}
\acrodef{SPE}[SPE]{Software Performance Engineering}
\acrodef{SPINA}[SPINA]{Superimposed Peer Infrastructure for iNformation Access}
\acrodef{SPLIT}[SPLIT]{Stemming Program for Language Independent Tasks}
\acrodef{SPOOL}[SPOOL]{Simultaneous Peripheral Operations On Line}
\acrodef{SQL}[SQL]{Structured Query Language}
\acrodef{SR}[SR]{Sliding Ratio}
\acrodef{sRBP}[sRBP]{Session Rank Biased Precision}
\acrodef{SRS}[SRS]{Simple Random Sampling}
\acrodef{SRSWOR}[SRSWOR]{Simple Random Sampling without Replacement}
\acrodef{SRSWR}[SRSWR]{Simple Random Sampling with Replacement}
\acrodef{SRU}[SRU]{Search/Retrieve via \acs{URL}}
\acrodef{SS}[SS]{Sum of Squares}
\acrodef{SSD}[s.s.d.]{statistically significantly different}
\acrodef{SSTF}[SSTF]{Shortest Seek Time First}
\acrodef{STAR}[STAR]{Steiner-Tree Approximation in Relationship graphs}
\acrodef{STELLA}[STELLA]{InfraSTructurEs for Living LAbs}
\acrodef{STON}[STON]{STemming ON}
\acrodef{SVM}[SVM]{Support Vector Machine}
\acrodef{TAC}[TAC]{Text Analysis Conference}
\acrodef{TBG}[TBG]{Time-Biased Gain}
\acrodef{TCP}[TCP]{Transmission Control Protocol}
\acrodef{TEL}[TEL]{The European Library}
\acrodef{TERRIER}[TERRIER]{TERabyte RetrIEveR}
\acrodef{TF}[TF]{Term Frequency}
\acrodef{TFR}[TFR]{True False Rate}
\acrodef{TLD}[TLD]{Top Level Domain}
\acrodef{TME}[TME]{Topics--Metrics-Experiments}
\acrodef{TN}[TN]{True Negative}
\acrodef{TO}[TO]{Transfer Object}
\acrodef{TP}[TP]{True Positve}
\acrodef{TPR}[TPR]{True Positive Rate}
\acrodef{TRAT}[TRAT]{Text Relevance Assessing Task}
\acrodef{TREC}[TREC]{Text REtrieval Conference}
\acrodef{TRECVID}[TRECVID]{TREC Video Retrieval Evaluation}
\acrodef{TTL}[TTL]{Time-To-Live}
\acrodef{UCD}[UCD]{Use Case Diagram}
\acrodef{UDC}[UDC]{Universal Decimal Classification}
\acrodef{uGAP}[uGAP]{User-oriented Graded Average Precision}
\acrodef{UI}[UI]{User Interface}
\acrodef{UML}[UML]{Unified Modeling Language}
\acrodef{UMT}[UMT]{User Management Tool}
\acrodef{UMTS}[UMTS]{Universal Mobile Telecommunication System}
\acrodef{UNESCO}[UNESCO]{United Nations Educational, Scientific and Cultural Organization}
\acrodef{UoM}[UoM]{Utility-oriented Measurement}
\acrodef{UPC}[UPC]{Universal Product Code}
\acrodef{URI}[URI]{Uniform Resource Identifier}
\acrodef{URL}[URL]{Uniform Resource Locator}
\acrodef{URN}[URN]{Uniform Resource Name}
\acrodef{USM}[USM]{User Storing Manager}
\acrodef{VA}[VA]{Visual Analytics}
\acrodef{VAIRE}[VAIR\"{E}]{Visual Analytics for Information Retrieval Evaluation}
\acrodef{VATE}[VATE$^2$]{Visual Analytics Tool for Experimental Evaluation}
\acrodef{VIRTUE}[VIRTUE]{Visual Information Retrieval Tool for Upfront Evaluation}
\acrodef{VD}[VD]{Virtual Document}
\acrodef{VDM}[VDM]{Visual Data Mining}
\acrodef{VIAF}[VIAF]{Virtual International Authority File}
\acrodef{VIM}[VIM]{International Vocabulary of Metrology}
\acrodef{VL}[VL]{Visual Language}
\acrodef{VoIP}[VoIP]{Voice over IP}
\acrodef{VS}[VS]{Visual Sentence}
\acrodef{W3C}[W3C]{World Wide Web Consortium}
\acrodef{WAN}[WAN]{Wide Area Network}
\acrodef{WHO}[WHO]{World Health Organization}
\acrodef{WLAN}[WLAN]{Wireless \acs{LAN}}
\acrodef{WP}[WP]{Work Package}
\acrodef{WS}[WS]{Web Services}
\acrodef{WSD}[WSD]{Word Sense Disambiguation}
\acrodef{WSDL}[WSDL]{Web Services Description Language}
\acrodef{WWW}[WWW]{World Wide Web}
\acrodef{XAI}[XAI]{eXplainable \acs{AI}}
\acrodef{XMI}[XMI]{\acs{XML} Metadata Interchange}
\acrodef{XML}[XML]{eXtensible Markup Language}
\acrodef{XPath}[XPath]{XML Path Language}
\acrodef{XSL}[XSL]{eXtensible Stylesheet Language}
\acrodef{XSL-FO}[XSL-FO]{\acs{XSL} Formatting Objects}
\acrodef{XSLT}[XSLT]{\acs{XSL} Transformations}
\acrodef{YAGO}[YAGO]{Yet Another Great Ontology}
\acrodef{YASS}[YASS]{Yet Another Suffix Stripper}

%% file: section/abstract.tex
\begin{abstract}
During the workshop, we deeply discussed what \acf{CONIAC} is and its unique features, proposing a world model abstracting it, and defined the \acf{CAFE} for the evaluation of \ac{CONIAC} systems, consisting of six major components: 1)~goals of the system's stakeholders, 2)~user tasks to be studied in the evaluation, 3)~aspects of the users carrying out the tasks, 4)~evaluation criteria to be considered, 5)~evaluation methodology to be applied, and 6)~measures for the quantitative criteria chosen. 
\end{abstract}

%% file: section/executive-summary.tex
\section*{Executive Summary}

This workshop brought together $22$~experts from both academia and industry in neighboring areas---namely \acf{IR}, \acf{RS}, and \acf{NLP}---in order to envision a new framework for the evaluation of \acf{CONIAC} systems and to discuss the related research challenges.

The framework starts from the assumption that a \ac{CONIAC} system will be able to \begin{enumerate*}[label=(\roman*)] 
\item \textit{interact} with users more naturally and seamlessly, \item guide a user through the process of \textit{refining} and \textit{clarifying} their needs,
\item \textit{aid decision-making} by making \textit{personalized recommendations and information} while being able to \textit{explain} them, and \item \textit{generate}, \textit{retrieve} and \textit{summarize} relevant information
\end{enumerate*}. To this end, a \ac{CONIAC} system needs to cross the boundaries and mix different technologies, e.g., \ac{IR} and \ac{RS}, rely on both internal knowledge about the user and external knowledge about the context where it is operating, and, finally, track the stream of conversation events and dynamic changes in the user's belief and information state.

To support the development of such \ac{CONIAC} systems, the \acf{CAFE} recognizes that, to follow the flow of conversational events and system states, evaluation needs to consist of a dynamic \emph{sequence of evaluation probes}, rather than being a single static assessment as it is today. Then \ac{CAFE} introduces six areas to guide researchers and developers in defining what the evaluation probes should be for the case at hand:
\begin{itemize}
    \item \textbf{Stakeholder Goals} discusses the diverse and often implicit objectives of various stakeholders involved in system design. These stakeholders include users, system designers, app developers, distribution platforms, content creators, publishers, advertisers, and editors. Each group has distinct goals that may overlap or conflict with others. For a system to be successful, it must address and balance these differing objectives. The section also notes the importance of enriching user interactions by supporting secondary goals, such as relationship-building and trust assessment, which enhance the overall user experience.
    
    \item \textbf{User Aspects} investigates the different angles of users approaching systems with diverse needs, such as answering questions, gaining knowledge, planning trips, or asking for recommendations. Some users or user contexts might benefit more from conversational systems due to some characteristics such as impairments or modality (e.g., hands-free operation). Users’ conversational information access tasks can also vary based on cultural contexts, affecting how they interact with systems. When evaluating the system, it will be essential to consider these user aspects, as different users may have very different conversations and experiences with the system. 

    \item \textbf{Task} explores the characteristics and tasks suitable for \ac{CONIAC} systems, which are defined by factors such as information need, human involvement, goal orientation, and iterative interaction. The proposed model categorizes tasks based on their complexity and definiteness, with well-defined tasks needing focused iterative interactions and ill-defined tasks requiring exploratory search and learning. This section presents examples, such as hypothesis formulation, product search, travel planning, and health information seeking, illustrating how \ac{CONIAC} systems can adapt to evolving user needs and provide more tailored assistance compared to conventional systems. The discussion extends to emerging LLM-based conversational enterprise and personal information management.

    \item \textbf{Criteria} discusses the criteria to consider when instantiating an evaluation framework for the conversational search scenario. Such criteria can be organized into a taxonomy according to who the subject of the evaluation is. The first layer of this taxonomy divides the criteria into system- and user-centric criteria. The former concerns the system and can be assessed in isolation in a purely offline setting. The role of the human, in this case, is to provide labels. The latter regards user experience and the consequences associated with the usage of the system. These top-level criteria are then further specialized along several dimensions. 

    \item \textbf{Methodology} organizes evaluation methodologies---both quantitative and qualitative---along two dimensions: (1)~according to the focus of a study (a standard dimension in \ac{IR}), ranging from system-focused methodologies like offline simulations to user-focused methodologies like qualitative interviews; and
(2)~according to the employed time model (a dimension from system behavior theory that is especially suitable for \ac{CONIAC} studies), ranging from stationary methodologies like single-interaction experiments to methodologies like longitudinal user studies that allow for continuous measurements of, for example, satisfaction.
Indeed, while evaluation criteria are defined as agnostic of time models, for the actual evaluation one has to decide which time model to use, which limits the available evaluation methodologies and affects the details of their implementation. 

    \item \textbf{Measures} presents the commonly used metrics for assessing system-centric hardware and software criteria, and the measures for evaluating user-centric criteria (including users' subjective reflection on their interactions, their perceptions, and behavioral metrics). Depending on the goals and criteria a specific system aims to fulfill, one can adopt the suitable evaluation methodology (see examples in \cref{tab:criteria-methods}) to assess the corresponding metrics/measures. 
\end{itemize}

When designing an evaluation, the first step is to identify the stakeholders and their goals that need to be addressed. Based on the goals, the user tasks to be studied in the evaluation have to be defined, and the user aspects need to be considered. The central elements of an evaluation are the criteria to be focused on, which can be determined by the stakeholder goals. The chosen criteria restrict the range of possible evaluation methods (e.g.\, any user-centric criterion requires the involvement of actual users in the evaluation procedure). Finally, an appropriate measure has to be defined for any quantitative criterion.

Overall, the \ac{CONIAC} world model and the \ac{CAFE} framework propose a series of ``\textit{high risk, high gain}'' research topics that promise to deliver a major paradigm shift in the field of conversational agents. By embracing a new vision for conversational information access and targeting a technological breakthrough, these initiatives aim to transform how academia and industry invent, design, and develop such systems.

%% file: section/introduction.tex
\section{Introduction}
\label{sec:introduction}

Every day, individuals interact with a wide variety of \textit{agents}: web search engines to find information, recommender systems to identify content or products to purchase, or chatbots for customer service inquiries. Given their prevalence, let us focus on \textit{conversational agents}, those that mimic human conversation using either text or spoken language, and more particularly on those enabling \textit{information access}. While the literature in this area is rich~\cite{ZamaniEtAl2023}, the evaluation of these systems is still limited to simplistic approaches that mostly ignore the conversational nature of the human-system interaction. In this manifesto, we specially focus on information access systems~\cite{ShahBender2024,zz-WhiteShah2025-editor} which allow us to seek, access, and use relevant information.

The very idea of some conversational interaction with information access systems traces back several decades. For example, \citet{Codd1974} early imagined conversations as a key way to allow casual users to interact with databases in a natural way, avoiding the complexity of a formal query language. Conversations figured out by \citeauthor{Codd1974} were very close to the idea of conversation we are still talking about nowadays, e.g., \citet[p.~182]{Codd1974}: 
\begin{verbatim}
    USER: Give me a list of the part numbers, quantities, 
          and suppliers of Houston parts 
    
    SYSTEM: You are requesting (among other thing) a listing of 
            quantities of parts. Do you want: 
            1. quantities on hand 
            2. quantities on order 
            3. quantities shipped?
    
    USER: I want quantities on order 
    
    SYSTEM: Fine! Now, with regard to "Houston parts" do you mean: 
            1. parts supplied by Houston supplier
            2. parts supplied to Houston projects
            3. parts associated with Houston in some other way?

    USER: I mean item 2
\end{verbatim}
The above conversation reveals the assumption of querying a relational database and the need to specify which relations/tables to query, possibly joining them, and which attributes within those relations/tables.

Recent years have witnessed a spurring of research on conversational agents, in multiple intersecting domains, such as \acf{AI}~\cite{GaoEtAl2019}, \acf{IS}~\cite{ZamaniEtAl2023}, \acf{IR}~\cite{KeyvanHuang2023,zz-DagstuhlSeminar19461,KieselEtAl2021}, \acf{RS}~\cite{JannachEtAl2022}, \acf{NLP}~\cite{RahmaniEtAl2023,LiangEtAl2023}, and even more with the rise of \acfp{LLM} and \acf{RAG}~\cite{AllanEtAl2024}.

A natural question arises after so many years of research by multiple disciplines: \emph{``Isn't this a solved problem?''}.

\begin{figure}[tb]
    \centering
    \includegraphics[width=0.9\linewidth]{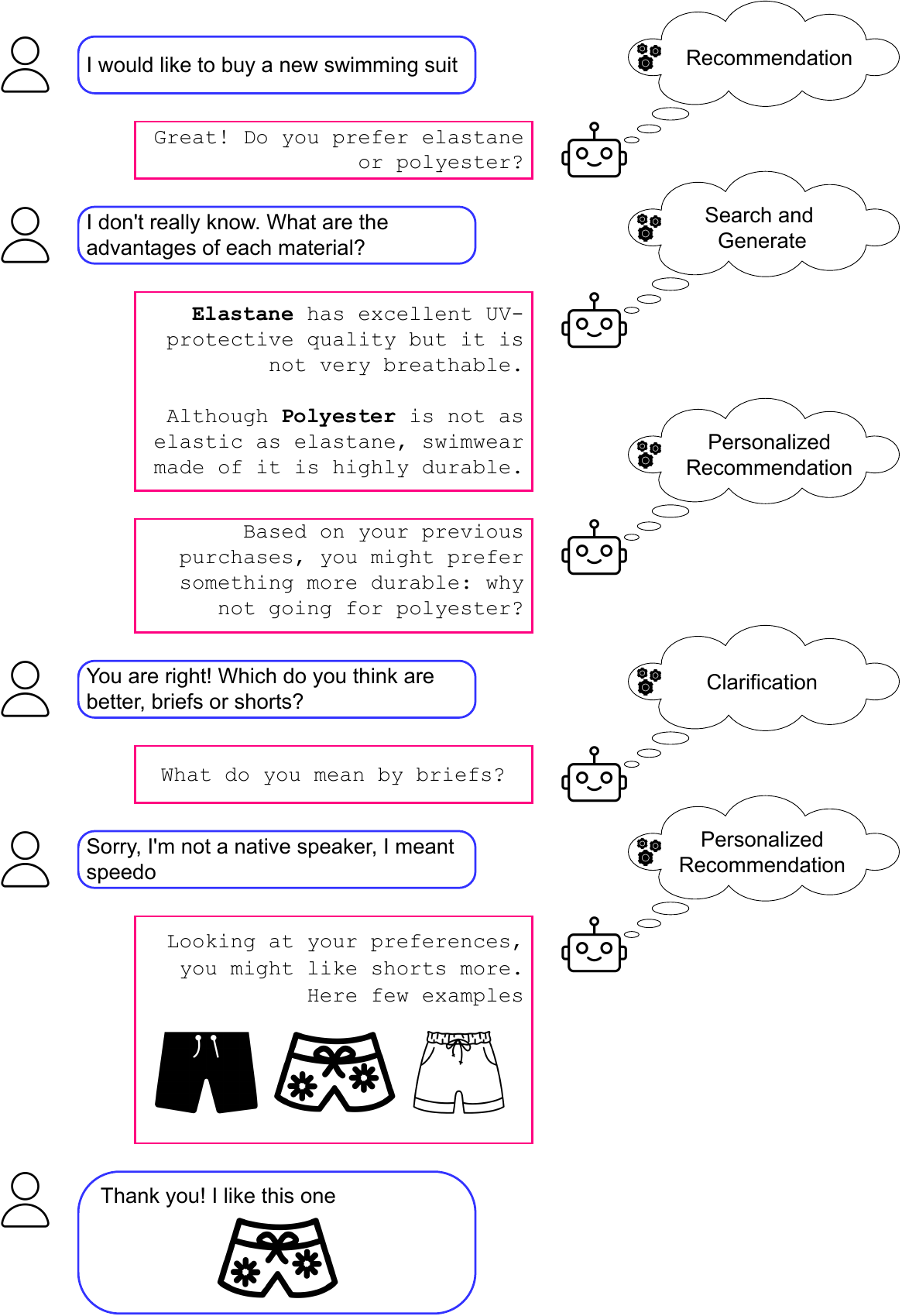}
    \caption{An example of a possible conversation in the product search domain.}
    \label{fig:conversation}
\end{figure}

In order to answer this question, in \cref{fig:conversation}, we present an example of how we imagine future conversations between humans and agents, inspired by \citet{DiNoiaEtAl2024}. We choose product search, a well-explored domain where it should be easier to answer ``yes, it is a solved problem!'' Differently from the early conversation imagined by \citet{Codd1974} and confined to the realm of relational databases, in our case, the conversation naturally spans different types of systems and methodologies---e.g., \ac{IR} systems and \acp{RS}---even if some commonalities are there, such as the system asking for clarifications when needed.

In the example, the user wants to buy a new swimsuit, but he is initially unsure or unaware of his exact requirements for this swimsuit. To support his decision-making process, the user initiates a dialog with the system: ``I would like to buy a new swimming suit.'' The system understands that the user's information need is best served by a \ac{RS} component. Therefore, it responds with an attribute-related question: ``\texttt{Great! Do you prefer elastane or polyester?}'' Since the user does not know the pros and cons of the two materials, he asks for an explanation of the system: ``I don't really know. What are the advantages of each material?'' Initially, the system relies on a combination of \ac{IR} and generation, e.g., a \ac{RAG} component, to answer the user information need; afterward, it relies on a personalized \ac{RS} to suggest going for polyester, since this material best matches with the user preference for more durable items. The user appreciates the suggestion and goes on asking ``Which do you think are better, briefs or shorts?'' Since the system does not correctly understand the term ``briefs'', it relies on a clarification component to ask for an explanation, ``\texttt{What do you mean by briefs?}'' Once the user clarifies that he actually means ``speedo'', the system relies on a personalized \ac{RS} component to suggest some possible shorts---a type of swimsuit that better fits with the user preferences and profile---and, finally, the user picks one of them, ending the conversation. \cref{fig:conversation} also shows that, while we typically think of conversations made up just of textual utterances, we should generally consider \emph{multimodal} and \emph{multimedia} conversations, where more complex phenomena may happen as, e.g., interruptions in case of a spoken interaction; in this simple example, this fact is represented by images of swimsuits in some utterances.

\begin{figure}[tb]
    \centering
    \includegraphics[width=0.78\linewidth]{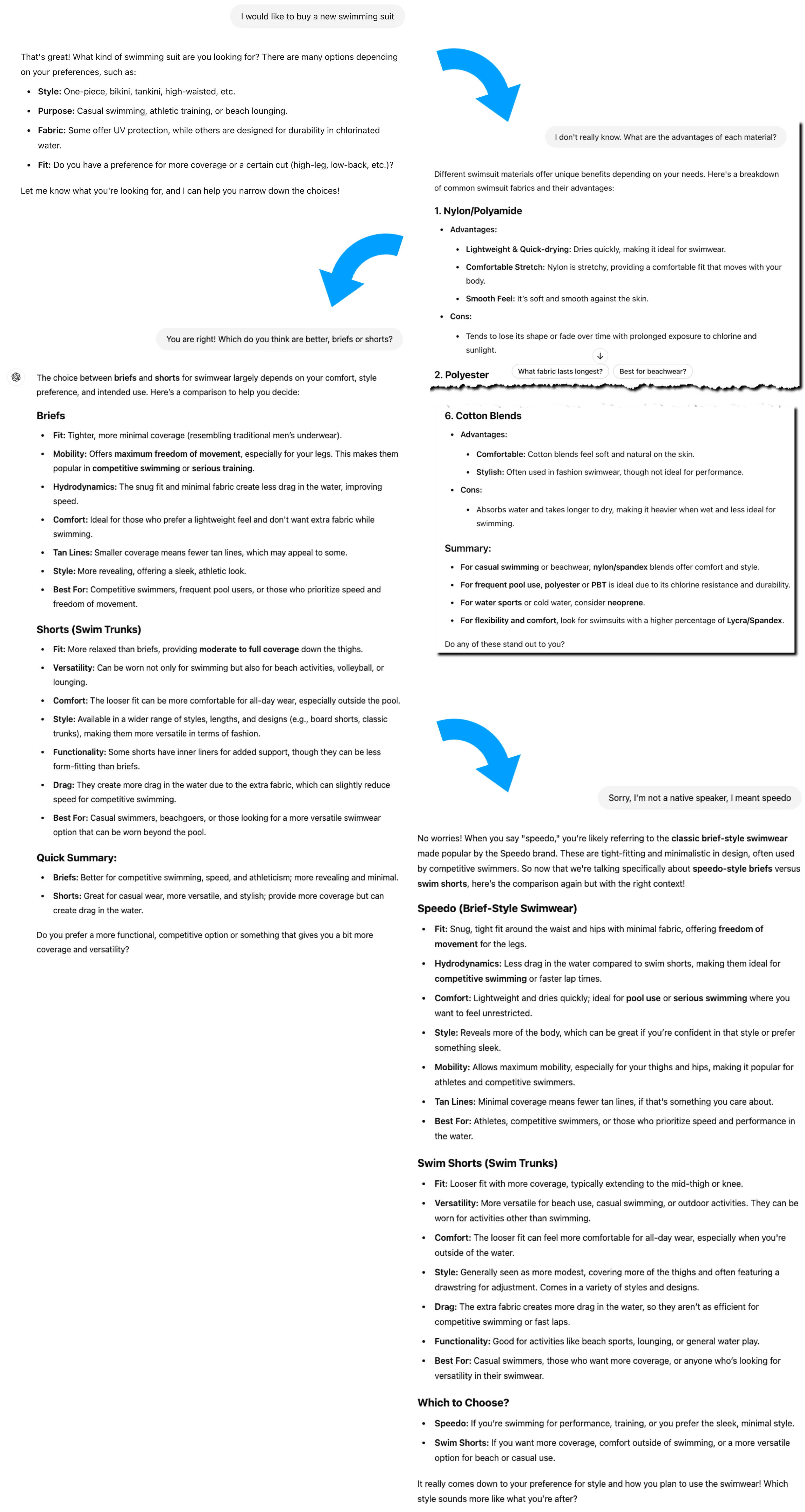}
    \caption{An example of conversation of \cref{fig:conversation} using GPT-3.}
    \label{fig:gpt-3}
\end{figure}

\begin{figure}[tb]
    \centering
    \includegraphics[width=0.99\linewidth]{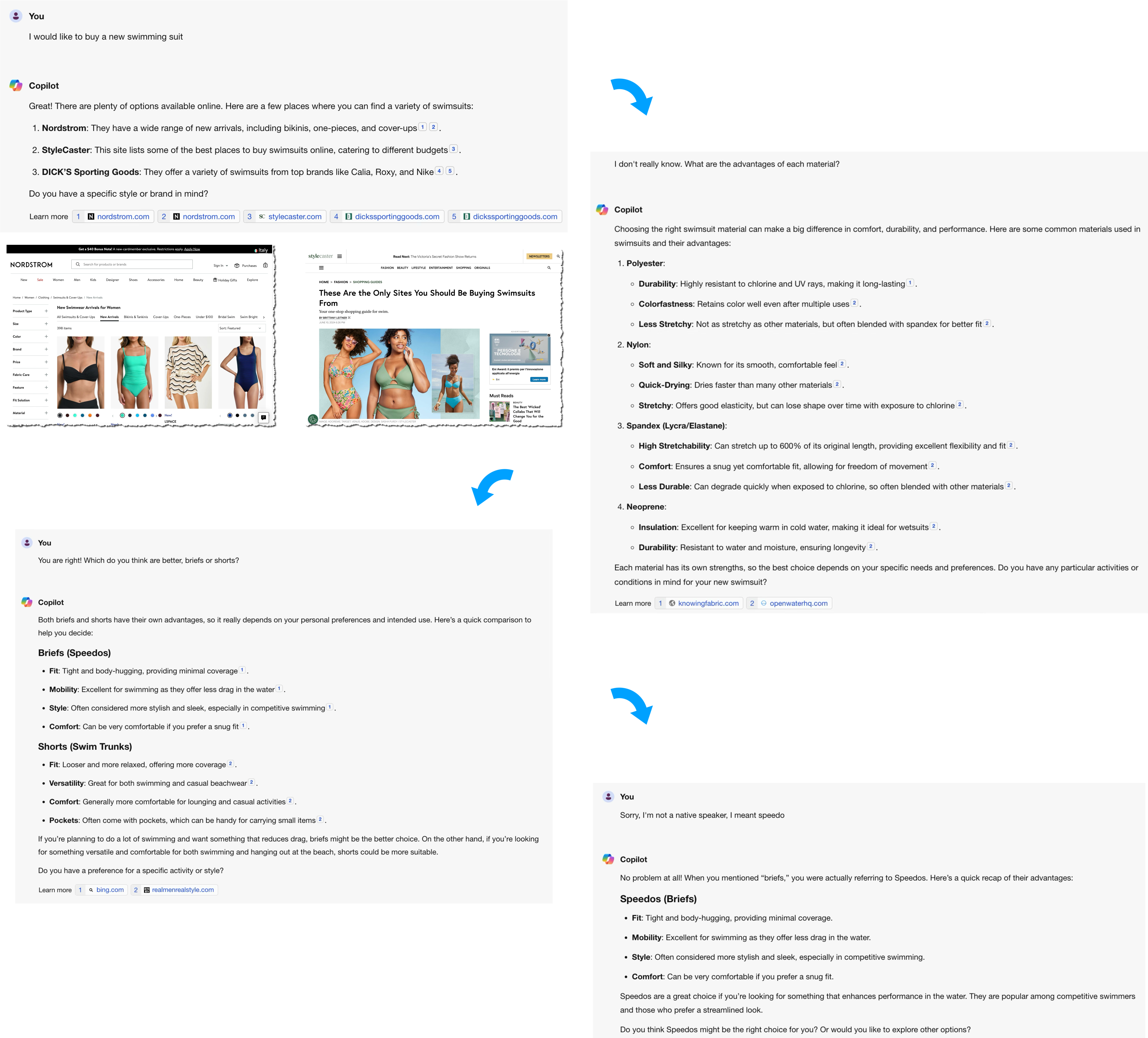}
    \caption{An example of conversation of \cref{fig:conversation} using GPT-4 via Microsoft Copilot.}
    \label{fig:gpt-4}
\end{figure}

To assess the extent to which this hypothetical conversation can be already served by current technology, we carried out a toy (and qualitative) experiment and ran this simplistic conversation, i.e., the sequence of utterances issued by the user of~\cref{fig:conversation}, using both GPT-3\footnote{\url{https://chatgpt.com/}}, shown in~\cref{fig:gpt-3}, and GPT-4 via Microsoft Co-pilot\footnote{\url{https://copilot.cloud.microsoft/}}, shown in \cref{fig:gpt-4}. The experiment was run in mid-October 2024 and, obviously, has several limitations, among which is the fact that GPT-3 and GPT-4 neither are backed by a \ac{RS} nor do these have a specific user profile that can be used for personalization. Nevertheless, the type of answers returned gives an impression of how much of the problem is solved.

The example conversation with GPT-3 in~\cref{fig:gpt-3} highlights some limitations. First, the generated answers tend to be quite lengthy, more than what a user might expect, and they are perhaps schematic, tending to rely a lot on item lists; on the positive side, utterances typically include a final recap. Obviously, GPT-3 cannot follow up the comparison among different materials with a personalized recommendation. GPT-3 does not perceive any ambiguity between \texttt{briefs} and \texttt{speedos}, while native speakers might do, and the clarification by the user produces a slightly more adapted response. On the positive side, GPT-3 maintains the context between the two utterances and continues to compare \texttt{shorts} and \texttt{speedos}. On the negative side, we can observe as in all the responses, the dimensions of comparisons are somewhat inconsistent since, for example, when it comes to materials they consist of entirely different facets or when comparing \texttt{briefs} vs.\ \texttt{shorts} and then \texttt{speedos} vs.\ \texttt{shorts} some explanations move from one category to another, e.g., freedom of movement is under mobility for \texttt{briefs} and under fit for \texttt{speedos}. Again, GPT-3 cannot complete the interaction with a personalized recommendation as in the case of~\cref{fig:conversation}.

The example conversation with GPT-4 in~\cref{fig:gpt-4} highlights some limitations. On the positive side, GPT-4 strives to provide references to support its answers. On the negative side, the answer to the first utterance is very much focused on just recommending shops or brands rather than exploring the actual information need of the user. Moreover, the answer itself is a bit mixed and not fully homogeneous because some of the suggestions point to e-commerce websites (e.g., Nordstrom) while others are fashion blog posts (e.g., StyleCaster); this somewhat recalls the issue mentioned above for GPT-3 about the dimensions of comparison being inconsistent. When it comes to comparing alternatives, the answers are much more compact and to the point than in the case of GPT-3, still suffering, however, from inconsistent dimensions across different materials. As with GPT-3, GPT-4 is not capable n of following up the explanation of materials with a personalized recommendation. When it comes to the comparison between \texttt{briefs}/\texttt{speedos} vs.\ \texttt{shorts}, GPT-4 explicitly considers them as synonyms, while a native speaker might not do so. If we force GPT-4 to answer just about \texttt{speedos}, on the positive side, the answer remains fully consistent with what replied in the case of \texttt{briefs}/\texttt{speedos} but, on the negative side, context is completely lost, and no comparison to \texttt{shorts} is made anymore. Also, GPT-4 does not complete the interaction with a personalized recommendation.

If we compare the interaction style of GPT-3 with that of GPT-4, apart from observing the more succinct answers in the latter case, we can also see that GPT-4 is very much focused on somehow narrowing down the scope of the conversation because the initial attempt is to immediately propose products, rather than exploring the actual needs of the stakeholder and asking for clarifications.

Overall, while the \acp{LLM} and chatbots of today are advanced systems, already performing well, they are currently insufficient to fully embody the vision (outlined above) of conversational interaction with information access systems, even in a well-studied and relatively confined domain such as product search and recommendation.

It is not hard to imagine more complex information access challenges, such as finding an object described by both structured data and free text elements. We could imagine someone asking a system to find a walk to do one evening. The system could consider walks within the user's driving, cycling, or public transport range and reply with suggestions of different levels of difficulty, length, and ascent. The searcher could then ask the system to find a subset of walks where reviews have described sunset vistas. Here, the system might draw such information from multiple reviews. The system could also incorporate elements of recommendation by looking at the searcher's walk history and employing some form of collaborative filtering to  identify the ideal evening walk further.

The ideal system is likely to be even more complex. When one uses a traditional web search engine like Google, the system is actually a constellation of search services, each with its own searching approach: web search retrieves authoritative content, news retrieves from high-quality sources with a bias towards recency, image search will consider image quality and size, etc. Upon receiving a query, the search engine chooses which blend of services will best serve the query. A conversational information access system will likely need to provide the same suite of information access services that one sees in traditional search, as well as a way to identify which service best serves a particular conversation.

Therefore, we argue that we need to imagine a new generation of systems---we call them \acf{CONIAC} systems and introduce them in~\cref{sec:coniac}---able to adopt a more holistic approach to conversational interaction. Above all, and this is the primary focus of this report, we need to envision a new evaluation framework---we call it \acf{CAFE}, we introduce it briefly in~\cref{sec:cafe}, and then expand on its details in sections~\ref{sec:goals} to~\ref{sec:measure}---capable of supporting and driving the design and development of \ac{CONIAC} systems.

%% file: section/coniac.tex
Generative \section{\acf{CONIAC}}
\label{sec:coniac}

\begin{figure}[tb]
    \centering
    \includegraphics[width=\linewidth]{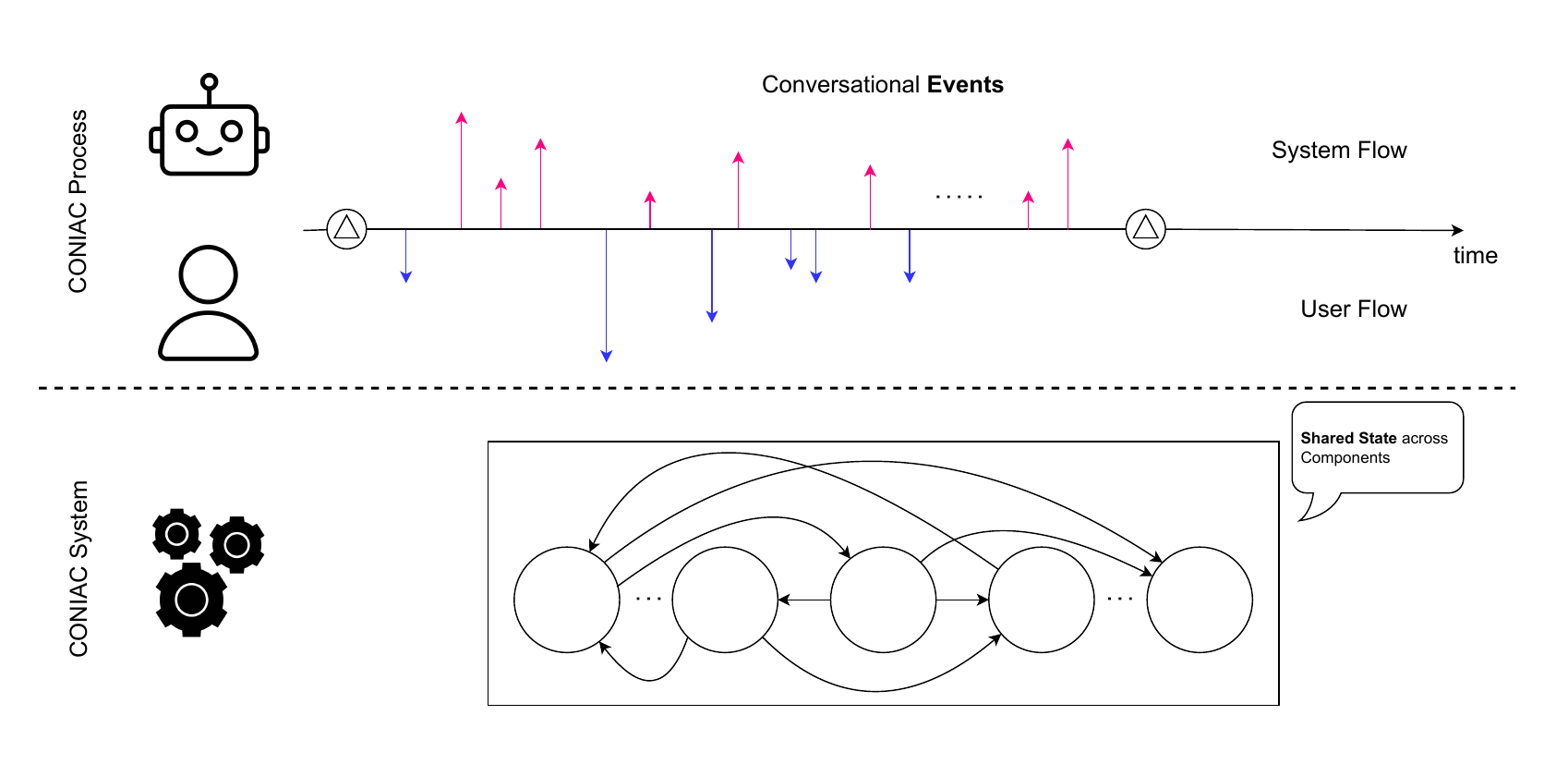}
    \caption{The \ac{CONIAC} World Model.}
    \label{fig:coniac}
\end{figure}

\cref{fig:coniac} depicts the \emph{\acf{CONIAC} World Model}, where we take a very broad view by considering any context and system that addresses a user's information need using language interactions. Indeed, given the wide range of domains and tasks that can prompt conversational events---each with its inherent peculiarities---it is impractical to exhaustively define every information discovery use case that might benefit from conversational aspects. Therefore, in the rest of this section, we introduce a layered abstraction---the \ac{CONIAC} World Model---that allows us to represent the conversational process targeting information access from various perspectives. 

Our broad view encompasses aspects of conversational information access systems as defined by~\citet{balog2021conversational}:
\begin{quote}
[W]e use the term conversational information access (CIA) to define a subset of conversational AI systems that specifically aim at a task-oriented sequence of exchanges to support multiple user goals, including search, recommendation, and exploratory information gathering; that require multi-step interactions over possibly multiple modalities. Further, these systems are expected to learn user preferences, personalize responses accordingly, and be capable of taking initiative.
\end{quote}
and conversational information seeking as defined by \citet{ZamaniEtAl2023}:
\begin{quote}
\ac{CIS} 
is concerned with
a sequence of interactions between one or more users and
an information system. Interactions in \ac{CIS} are primarily
based on natural language dialogue, while they may include
other types of interactions, such as click, touch, and body gestures. 
\end{quote}

While we take a broad view of \ac{CONIAC} systems, one that encompasses both traditional conversational \ac{IR} and \ac{RS}, we envision that \ac{CONIAC} will be most useful in interactive scenarios when users have complex or poorly understood information needs.

The \ac{CONIAC} World Model consists of two layers: 
the \ac{CONIAC} Process Layer (\cref{ssec:processlayer}), where the actual conversation happens, and 
the \ac{CONIAC} System Layer (\cref{ssec:systemslayer}), which is what we aim to design and develop.

\subsection{The \acs{CONIAC} Process Layer}
\label{ssec:processlayer}

We use the term \emph{\ac{CONIAC} Process Layer}, or \textbf{conversational process} for short, to refer to the overall conversation flow, i.e. any interaction between a \ac{CONIAC} agent (the System Flow) and a human (the User Flow), via the interface of the system in a \emph{multimedia} and \emph{multimodal} way, in the most general setting. Some examples of conversational processes and mechanisms that influence the actual moves, utterances, or turns being generated are those that establish and maintain interpersonal rapport; those that organize the interaction channel through turn-taking cues, meta-communicative turns and even non-verbal actions such as gestures, facial expressions, and laughter; those that structure the informational flow of the conversation and narration as it proceeds, manage continuity and coherence, and negotiate topical shifts or switches when necessary; and those that define, modify, and address discourse referents and their properties. These include turns such as plain utterances, clarification questions, greetings, repairs, and verifications~\cite{hobbs1979coherence,grosz1986attention,hobbs2014towards,trippas2020towards}. Many conversational mechanisms are intuitively acquired by humans without instruction, though some are more explicitly trained in education or professional contexts.

More specifically, and more related to information access, conversational processes of interest include system and user collaborating in the expression of an information need, revealing the system's capabilities and collection qualities, clarification questions using mixed initiative, and persistence of discourse referents, allowing both parties to refer back to recently mentioned items or sets; and the ability to reason about the qualities and utility of a set and to further refine it for the purpose at hand~\citep{radlinski2017theoretical}.

An important assumption we make is that every conversational process addresses a single information need and thus has a single objective. This is a necessary assumption for our framework as, later on, this objective informs the selection of aspects to be evaluated, e.g., which criteria or measurements to use.

Specifically, a conversational process comprises different \textbf{conversational events} that occur over continuous time. We mean event in the most abstract sense, ranging from simple and atomic textual utterances to complex and \emph{compound multimedia} and \emph{multimodal} interactions, where phenomena such as interruptions, which for instance is central to spoken interaction, may happen. We distinguish four main conversational events---start, end, user, and system events---, that will be described in greater detail in the following paragraphs.

We further assume that an arbitrary amount of time may pass between two conversational events. Importantly, this includes possible interruptions, breaks, or pauses, after which the user or system decides to resume the conversation. This requires that the ``state'' of the conversation is maintained for arbitrary lengths of time. While concrete implementations may differ concerning the duration for which the state of the conversational process is preserved, we assume here that users may abandon and return to a session at any time without loss of generality. 

Firstly, we make explicit that every conversational process starts with a \textbf{start event} and ends with a \textbf{end event}. Between this start and end event, the conversational process is comprised of ``system'' and ``user'' events, both of which correspond to language (or more general) interactions, either issued by the system (system events) or the user (user events). Importantly, we do not require that conversational processes alternate between user and system events. Users may issue several ``user events'' in a row, and so may the system. This is exemplified, for instance, in~\cref{fig:conversation} where in response to the user event, ``I don't really know. What are the advantages of each material?'', the system issues two events, one explaining the advantages of each material and the other to perform a personalized recommendation.

Importantly, we distinguish start and end events from other types of conversational events. This is because the start of the conversational process can be a user event---e.g., ``I need to learn about Hillary Clinton's career as a Secretary of State for a test. What do I need to know?''---or a system event---``Hello, my name is Sven. I am here to help you. What would you like me to help you with?''---, but also other types of non-language interactions between the user and the system, e.g., the user opening the chat window. The same is true of the end event. 

Despite its apparent simplicity, this abstract and event-based view of the conversational process is flexible and powerful enough to be an instantiated model and represents all the conversational mechanisms and information discovery use cases briefly exemplified above.

\subsection{The CONIAC System Layer}
\label{ssec:systemslayer} 

Unlike general interactive information systems, a \ac{CONIAC} system must deal with the conversational dynamics described above, in particular by taking or relinquishing the initiative when necessary to keep the conversation flowing naturally~\cite{trippas2020towards}. In this regard, \ac{CONIAC} systems differ from many other systems because the latter often have one-shot interactions or repeated interactions that can be viewed in isolation.
An example of the latter could be repeated query reformulation when using a search engine, e.g., ``Hillary Clinton'', followed by ``Hillary Clinton state secretary'', where each time the \ac{IR} system responds with a list of documents relevant to the current query and there is no notion of ``state'' from one query to the next. 
In contrast, in our process-oriented view of \ac{CONIAC} systems, this notion of \emph{state} is essential, as it allows the system to use knowledge of past interactions to refine current and future interactions.

The bottom part of \cref{fig:coniac} depicts the \emph{\ac{CONIAC} System Layer}. We make abstraction of real-world \ac{CONIAC} systems and only assume that the system will emit events that are generated by one or more components of the system, e.g., \ac{IR} or \ac{RS} components, essentially blurring the boundaries among components. These events can be, for example, responses to information needs, suggested items, explanations, or requests for clarification from the user. Note that by \ac{CONIAC} system, we refer to both the underlying application and its interface.

As mentioned previously, another important assumption we make as a result of our view of a conversation as a process requires that there is some \emph{shared state} across interactions between the user and the system, but also between different components---\ac{IR}, \ac{RS}, \ac{QA}---that together make up a \ac{CONIAC} system. 
A state may even be recorded across different sessions---conversational processes---with the system, resulting in personalized interactions tailored to the user's (interaction) preferences.

The system layer is critical for managing and guiding interactions in the conversation process layer. We envision the system layer as implementing a stateful machine capable of flexible/expressive conversation flow tracking, personalization, and information access functionalities. 
The following functionalities are, in our view, essential for handling complex tasks involving information access:

\begin{itemize}
    \item \textbf{World knowledge.} Reflects the external environment or domain-specific knowledge the system operates within, enabling it to provide contextually appropriate responses.

    \item \textbf{User information.} Incorporates user preferences or utilities, goals, and potentially private user-specific information, facilitating personalized interactions. 

    \item \textbf{State tracking.} This state tracking involves the context that tracks the flow (both user and system events defined in \cref{ssec:processlayer}) and the context of the conversation, including dynamically updated beliefs in the user information state revealed by user events.
\end{itemize}

State tracking in this scenario ensures that the \ac{CONIAC} system can proactively generate system events---such as recommendations, clarification questions, or acknowledgments---in a manner that is informed by both the ongoing conversation and the user's inferred goals and preferences. Such system events should be generated in light of a range of objectives---e.g., coherence and relevance throughout the conversation, task completion, and efficiency---as specified by the system designer.

As an example of a possible concrete instantiation of a \ac{CONIAC} system, \citet{boutilier2002pomdp} defines a preference elicitation recommendation conversation as a \ac{POMDP} model, which additionally requires a \emph{reward model} to encapsulate the agent's interaction objectives. One advantage of such a decision-theoretic system model (which may be optimized through reinforcement learning methods when exact model details are unknown) is that it provides algorithmic and learning methodologies for optimizing system actions concerning the system design. This stands in contrast to many existing conversational dialog systems leveraging rule-based decision-making that may not optimize an explicitly defined system objective.

\subsection{Limitations of Current Systems}

\subsubsection*{Limitations of \acl{IR} Systems.} 

Traditional \ac{IR} systems are particularly limited in their capacity to handle these scenarios because although the queries are ambiguous, the assumption in most cases is that users' information needs are relatively well-defined from the outset. 
These systems are primarily designed to interpret and respond to explicit queries, focusing on accurately understanding and estimating a pre-existing, well-formed information need~\citep{chang2020query,carpineto2012survey:query:expansion,anand2023queryunderstandingagelarge}. 
Consequently, they are less effective in situations when a user's needs are still evolving or are only partially articulated, which is common in many conversational contexts~\cite{DBLP:conf/chiir/KieselBSAH19}. 
In such scenarios, the rigid query-response paradigm of conventional IR systems falls short, as it does not adequately support the dynamic, iterative process of refining and clarifying user intent that is essential for effective conversational information access. 

\ac{IIR} systems are designed to support and even encourage user exploration within the information-seeking process, yet they are fundamentally limited by the fact that interaction is always initiated by the user. In these systems, the system consistently assumes the role of a passive responder, reacting to user queries rather than actively contributing to the dialogue. While such systems can effectively address well-defined user intents, they often fall short in scenarios where the user may have a clear goal but lacks awareness of all relevant aspects, possibilities, and available design or choice options~\citep{white2009exploratory,Marchionini2006exploratory}. 

This limitation means that users might not fully explore the breadth of information or decision-making options available to them, as the system does not proactively guide them toward uncovering unknown preferences or considerations. As a result, the interaction remains user-driven, potentially leading to missed opportunities for discovery and a less comprehensive understanding of the subject matter. To overcome these challenges, more advanced systems that can take on a more active role in the interaction, such as by prompting users with relevant questions, suggesting alternative pathways, or highlighting overlooked options, are needed. These systems would better support users in navigating complex information landscapes and making more informed decisions~\citep{DBLP:journals/ftir/Kelly09,Belkin2000_helpingpeople,white2009exploratory,Marchionini2006exploratory}.

\subsubsection*{Limitations of \acl{RS}.}
\acf{RS}, while highly effective in providing personalized content and suggestions based on user preferences, are also limited when applied to conversational scenarios. 
RS are inherently push-based systems, where suggestions are proactively presented to the user based on a pre-existing model of user behavior, preferences, and past interactions.
Although they are beneficial in providing insights about choices, aspects, and actions that are absent in \ac{IR} systems, recommender systems often lack query understanding capabilities. 
They do not interactively interpret or refine user intent based on evolving dialogue. 
Finally, both the \ac{IR} and the \ac{RS} approaches are mostly one-shot in the sense that they are optimized for single interaction settings. 
Even in session-based search~\citep{dalton2020treccast2019conversational} reformulations to an initial query result to a one-shot retrieval at the end of a session, and do not allow for intent modification or asynchronous change in intent.

%% file: section/cafe.tex
\section{\acf{CAFE}}
\label{sec:cafe}

In the world model depicted in \cref{fig:coniac}, we outline a generic representation of the conversational scenario. To evaluate in this context, we need to understand how and when measurements are made to validate the goodness of the system that guides information discovery. Goodness can be a complex concept to capture in a single measure for evaluation, as many facets can directly influence it. Consequently, we argue that depending on the use cases, \ac{CONIAC} systems are expected to encompass a wide range of different objectives and criteria for various stakeholders. 

Therefore, in order to explain the overall \acl{CAFE}, we first introduce the notion of an \emph{Evaluation Layer} (\cref{ssec:evallayer}) in the \ac{CONIAC} world model and, then, we explain of \ac{CAFE} (\cref{ssec:evalframework}) allows for a concrete instantiation of such layer for actual evaluation purposes.

\subsection{The Evaluation Layer}
\label{ssec:evallayer} 

\begin{figure}[tb]
    \centering
    \includegraphics[width=0.85\linewidth]{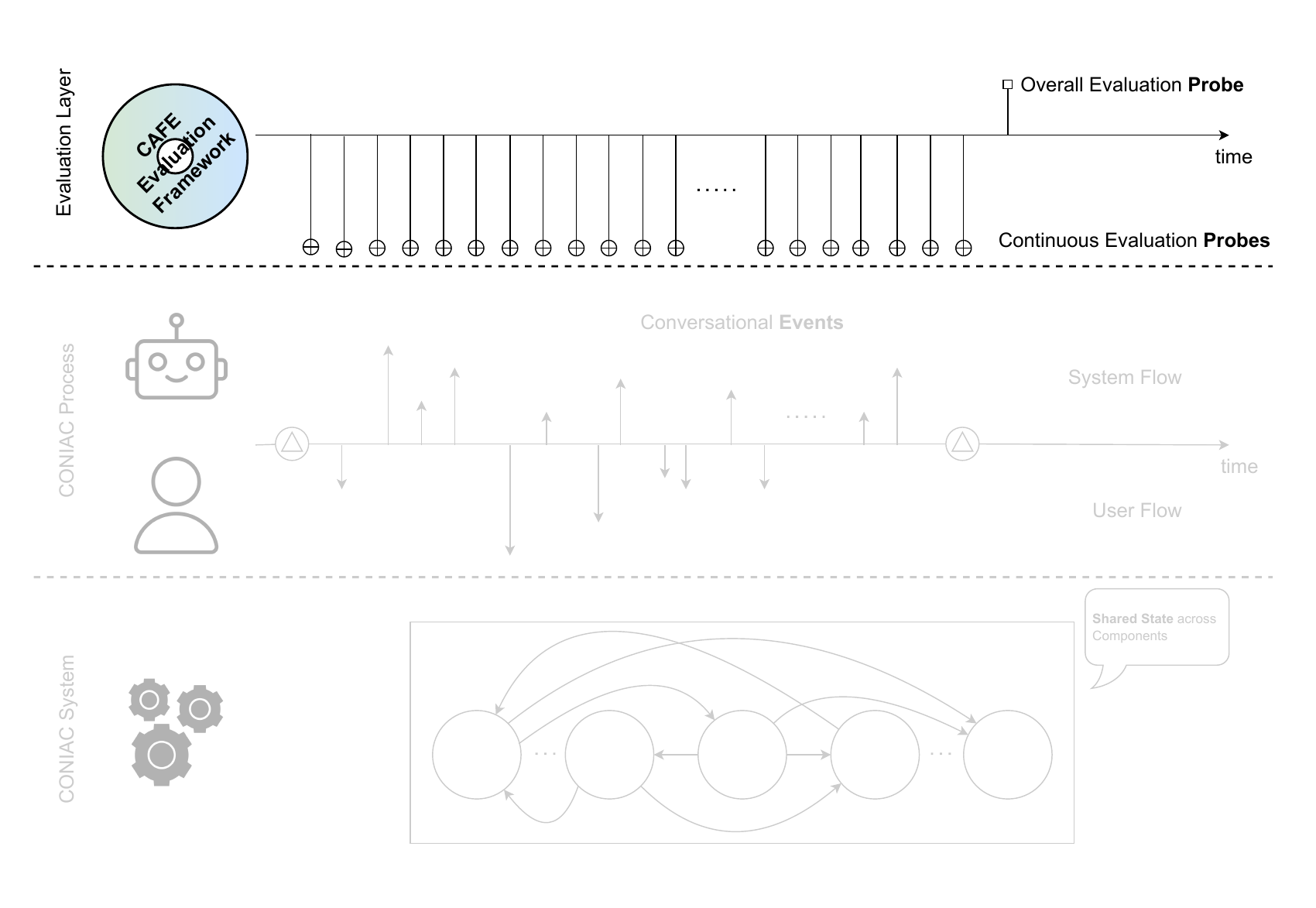}
    \caption{Evaluation Layer in the \ac{CONIAC} World Model.}
    \label{fig:eval-layer}
\end{figure}

The vision of the \ac{CONIAC} World Model depicted in the previous section calls for a paradigm shift in evaluation. As shown in~\cref{fig:eval-layer}, we can imagine that the \emph{\ac{CONIAC} Evaluation Layer} sits on top of the \ac{CONIAC} process and system layers, to allow for their assessment, where the \ac{CONIAC} system is the \emph{construct} under evaluation.

The literature is rich with examples of how to approach the evaluation problem from multiple perspectives~\citep{landoni2019sonny,liu2021meta, jannach2023evaluating,lipani2021doing,penha2020challenges}. However, in general, these assessment strategies focus on a more static approach, focused mostly on overall performance with the conversational system or user satisfaction with (their interactions with) said system. 

We argue that when it comes to \ac{CONIAC}, evaluation should be inherently \textbf{dynamic} and \textbf{multi-faceted}. Indeed, we are aware that there is no optimal set of measurements and measures that can, for every \ac{CONIAC} use case, comprehensively monitor the system (\cref{ssec:systemslayer}), the users (\cref{ssec:processlayer}), and the constantly evolving exchanges that take place. 
Consider an example of a direct answer to an unambiguous question `\textit{what is the weather like}'.
This direct answer might be assessed in terms of fast-time response and correctness (e.g., the system can provide the right information for the right location).
In contrast, for a more complex inquiry (e.g., a child explores to understand better a new concept (`\textit{Can animals sleep upside down}'), assessment should be more focused not just on the ability of the system to understand the information needed and provide a relevant response, but also on other perspectives inherent to this particular user. In this example, perspectives could include the ability of the system to provide suitable information and to respond in a manner that a child can comprehend \citep{landoni2019sonny,allen2023multi}. However, a clear pattern emerges: regardless of `what' (i.e., criteria) is assessed or `how' (i.e., measures), assessment of \ac{CONIAC} systems cannot be done in a summative manner, based on a single measurement.

With that in mind, we envision an evaluation layer that allows simultaneous and \textbf{continuous probing} of users, systems, and their interactions from multiple perspectives. Note that probes can happen synchronously with or asynchronously from the events in the conversational process, allowing for a wide range of evaluation styles.

In this layer, we define \textbf{probes} that act as sensors that constantly capture snapshots of the actions and reactions inherent to the conversational process from the start of the conversation all the way to the end (regardless of the resolution). Probes can be as simple as a single-valued measurement or complex multimedia objects, e.g., multi-valued measurements together with log records and video recordings of the user interaction. The idea is not to rely on a single probe, but instead---depending upon the use case---use a variety of probes that capture the different facets of the conversational process to assess.

The continuous probing is what then makes it possible not only to yield an overall measurement for the system---it can be some form of aggregation of the collected probes or an independent probe on its own---but also to gauge the correctness (or lack thereof) of the system events, along with user events. The measurements captured by these probes can also yield cumulative measurements, in addition to providing snapshots of the behavior of the different components of the system (and users' reactions to those).

Given that the probes can encode a rich and diverse set of measurements, it is possible to simultaneously capture measurements related to effectiveness and user behavior (via, for example, eye tracking), timeliness, and more. 

Note that we focused our discussion on how evaluation happens as a form of external assessment of the system. However, nothing prevents us from using the very same approaches to prompt system adaptation and correct the system behavior as it runs.

In the end, this layer can be instantiated based on a wide range of criteria for assessment for different \ac{CONIAC} systems, enabling the discovery of information in varied contexts and tasks. This is precisely the purpose of \ac{CAFE}, introduced in the next section.

\subsection{Evaluation Framework}
\label{ssec:evalframework} 

\begin{figure}[tb]
    \centering
    \includegraphics[width=0.85\linewidth]{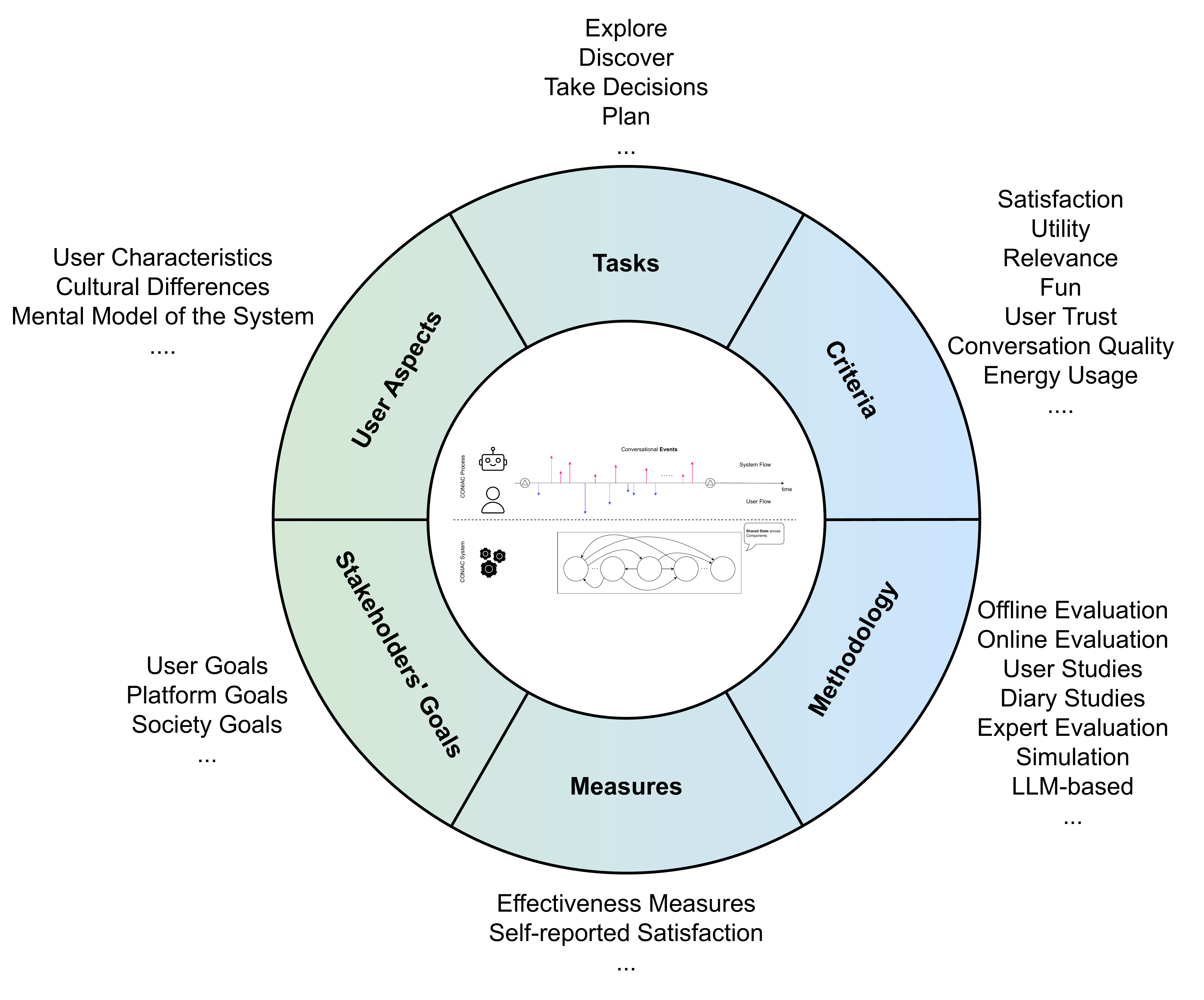}
    \caption{Evaluation Framework.}
    \label{fig:cafe}
\end{figure}

\cref{fig:cafe} illustrates our evaluation framework that accounts for the diverse sets of choices of tasks, stakeholder goals, and criteria that need to be taken to evaluate \ac{CONIAC} systems. We believe that our evaluation framework can not only cover most of the commonly used \ac{CONIAC} evaluation schemes but can guide a system designer to develop novel evaluation schemes based on both traditional and non-traditional measurements.

The main components of our framework account for different user aspects, stakeholder goals, tasks, criteria, evaluation methodologies, and measures. Following our evaluation framework, a system designer interested in evaluating a \ac{CONIAC} system should be able to develop an evaluation scheme that can be used as a continuous or multiple-shot experimental/evaluation probe as discussed in \cref{ssec:evallayer}.

\begin{itemize}

\item
    \textbf{Stakeholder goals.} Stakeholders of a \ac{CONIAC} system may have diverse goals that might or might not be directly accessible to system designers or evaluators and must often be implicitly inferred in evaluation. 
    \ac{CONIAC} systems might also have multiple goals ranging from end users having (in-)direct information needs to platforms deploying \ac{CONIAC} systems interested in content usage, user engagement, impression generation, and user retention, to name a few.

\item 
    \textbf{User aspects.} When developing an evaluation framework for \ac{CONIAC} systems, it is crucial to consider user-specific aspects, such as preferences, specialized needs, expertise types, and background characteristics, which may make conversational systems more beneficial than non-conversational alternatives.

\item
\textbf{Tasks.} \ac{CONIAC} involves tasks characterized by an information need, human involvement, goal orientation, and mixed-initiative between the user and the system.
While some tasks benefit from the introduction of a conversationally competent system, others may not, depending on the complexity of the task.

\item
\textbf{Criteria.} The scope of evaluation can range from single-turn interactions to entire conversations and long-term system usage, each requiring different criteria for assessment. Additionally, the temporal dimension, which examines how the system's performance changes over time, is a critical factor that intersects with both stationary and dynamic properties, making it essential for a comprehensive evaluation.

\item
\textbf{Methodology.}
In addition to the standard distinction of user-focused and system-focused methodologies, our evaluation framework categorizes evaluation methodologies also according to the employed time model---a dimension especially relevant for \ac{CONIAC}. This dimension ranges from stationary methodologies like single-interaction experiments to methodologies like controlled lab studies that allow for continuous measurements, such as physiological ones.

\item 
\textbf{Measures.} Finally, we allow for measures that typically focus on the system's ability to provide accurate, relevant, and timely information during interactions. 
Measures include objective measures of effectiveness to subjective notions of user satisfaction like self-reported satisfaction.
By incorporating both objective effectiveness measures and subjective self-reported satisfaction, evaluators can better understand the system's strengths and areas for improvement.

\end{itemize}

We offer details for each of the items above in Section ~\ref{sec:goals} to Section~\ref {sec:measure}.

We also take \cref{fig:cafe} as the \emph{workflow} to be adopted to evaluate \ac{CONIAC} systems. We start from defining the ``Stakeholders' Goals'' and progress clock-wise to ``User Aspects'', ''Tasks', ''Criteria'', ''Methodology'', and ''Measures''. All together, `Stakeholders' Goals'', ``User Aspects'', and ''Tasks' constitute the context in which the \ac{CONIAC} system is evaluated, whereas  ''Criteria'', ''Methodology'', and ''Measures'' define how the actual evaluation is to be carried out.

\subsection{Limitations of Current Conversational Evaluation}

The literature is rich in algorithms and strategies for handling multiple scenarios from conversation information access. While most of them are inspired by conversational search and conversational \ac{IR} literature, we have seen an interest in conversational recommendations over the last few years. Going through these works, a key commonality emerges: the lack of consensus on evaluating a conversational system in a manner that gives a holistic view of its performance and enables comparing and contrasting with other emerging technologies. Indeed, TREC tracks like TREC~CAsT~\cite{dalton2020treccast2019conversational,DaltonEtAl2020,dalton:2021,owoicho:2022} and TREC~iKAT~\cite{aliannejadi:2023,DBLP:conf/sigir/AliannejadiAC0A24} offer uniformity to comparisons for conversational search, but do so primarily with the relevance-driven focus for evaluation. Datasets such as ReDial~\citep{ReDial}, ConveRSE~\citep{ConveRSE}, or LLM-REDIAL~\citep{liang-etal-2024-llm} enable assessment of conversation for recommendations. However, as we previously stated (and illustrated with the earlier example), we emphasize that \ac{CONIAC} is a complex endeavor and
should be evaluated 
from multiple lenses~\citep{zangerle2022_fevr}. 

%% file: section/goal.tex
\section{Stakeholder Goals}
\label{sec:goals}

Stakeholders have varying goals, which are not generally accessible to system designers or evaluators and are not always explicitly formulated even for the stakeholder. They can often be inferred from the actions and the background information the system has available. Stakeholder goals are not always aligned: a system may benefit some stakeholders more than others \cite{surer2018multistakeholder,ekstrand2024not}. In general, stakeholders may have several simultaneous, interacting, and only partially overlapping goals~\citep{INR-079,bauer2019_multimethod_multistakeholder}. A conversational approach can further many stakeholder goals.

\begin{itemize}
    \item 
\textit{Users} have goals they wish to see a system help them achieve. A user may be solving an immediate problem or a long-term issue; wish to retrieve something they already know about (i.e., known item search); seek to learn or investigate a topical or thematic area \citep{marchionini2006}; they may be looking for diversion or entertainment or delight. Users may be focused on a goal, or the goal may be in the background for some other activity, and the goal may be of varying importance and centrality to the user throughout the task they are pursuing. Users may approach information access systems to find immediate factual responses, seek support for an argument, or plan a later investigation into an area. They may be seeking documents or collections of documents, they may be looking for some factual content they know or believe to be found in a document, or they may be looking to enjoy reading, viewing, or listening to some content that may be found in an item in a collection or across several items. To achieve these goals, users will generally want to expend little effort, enjoy interacting with the system, trust the correctness of the result, and feel confident that they have exhausted the usefulness of the system they are working with for the purposes they have engaged with it. 

\item 
\textit{Indirect users} are affected by the decisions and actions of the (primary) users based on the information they gained from the system. Examples are a doctor's patients consulting medical information systems or clients of lawyers who use their legal information systems. While the primary users may experience only gradual changes in the quality of their work due to the strengths and weaknesses of the information system, their individual decisions may have serious consequences for their clients.

    \item 
\textit{System designers} and \textit{App designers} wish to verify that their design is appealing, appropriate, effective, efficient, and catchy. 

    \item 
\textit{Distribution platforms} wish to retain the interest and appreciation of users, creators, and advertisers.

    \item 
\textit{Creators} provide materials they would like to see being used and enjoyed, and in many cases, they wish to receive remuneration for that usage.

    \item 
\textit{Publishers}, aligned with creators, wish to see their catalog prominently featured among the offerings to the users. 

    \item 
\textit{Advertisers} wish to receive many impressions for their messaging and to see those impressions convert to business opportunities.

    \item 
\textit{Editors and aggregators} wish to guide a user to further material.

    \item 
\textit{Other stakeholders can be added here}: surrounding agents such as colleagues, friends, family, or teachers.
\end{itemize}

In the following, we mainly focused on describing user goals, not on the other stakeholders' ones. However, it should be noted here that the functions of a system must satisfy, or at a minimum address, the goals of all stakeholders to be successful.

\subsubsection*{Top-level goals}

As deeply discussed in many venues~\citep{hleg2019ethics}, relevance, truthfulness, and trustworthiness\footnote{https://digital-strategy.ec.europa.eu/en/library/ethics-guidelines-trustworthy-ai} are primary and very common top-level goals. These top-level goals are typically well-represented and adequately considered in many evaluation activities in the field. If you look at the tasks offered by initiatives such as \ac{TREC}~\citep{zz-HarmanVoorhees2005-editor}, \ac{CLEF}~\citep{zz-FerroPeters2019-editor,Ferro2024}, \ac{NTCIR}~\citep{zz-SakaiEtAl2020-editor}, or \ac{FIRE}, you will find many evaluation efforts revolving around these top-level goals.

\subsubsection*{Minimizing effort}
Users wish to minimize the effort they expend on using a system. This means the interaction must have low friction and a low threshold. A conversational system will allow users to approach the system with little preparation and low cognitive load: they do not need to formulate their needs, tasks, and goals in terms exactly acceptable to the system since this can be negotiated during the session~\cite{trippas2020towards}. Conversely, using a conversational system may incur overhead---instead of immediate access to a known item through a short command, a user may need to engage in a more verbose interaction. A system needs to be convenient to the frequent power user and forgiving to the novice and the occasional user. The latter has more to gain from a conversational approach. 

The need for conversational systems to cater to power users and novices implies that evaluation metrics must consider varying user expertise and interaction styles, balancing efficiency and user satisfaction. Evaluators should assess systems based on their ability to minimize cognitive load and friction while providing flexible, adaptive interactions that accommodate user needs and preferences.

\subsubsection*{Establishing, clarifying, and elaborating objectives}

Conversational interaction allows the system and user to disambiguate ambiguous or polysemous concepts as well as clarify and specify terminology jointly. 

Conversational interaction allows systems and users to jointly explore and modify the shared understanding of some topic or theme of interest. By iterating over formulations, users can learn about the representation and terminology used in the system and of the topical coverage of a collection or a system; systems can gain a better understanding of user background and users' previous understanding of the topic or theme at hand. 
The ability of conversational systems to facilitate dynamic, interactive learning and exploration means that evaluation metrics must go beyond static measures of accuracy or relevance, encompassing the system's adaptability and capacity to refine user understanding over time.

\subsubsection*{Evolution and specialization of objectives}

Conversational interaction allows users to incrementally build on previous exchanges naturally and intuitively. This approach facilitates the user's exploration of a topic or theme and enables the system to offer relevant suggestions and guide the user better.

This allows users to \textit{specialize} their goals over the course of a conversation, starting with a general exploration of a topical area or a theme and then progressively drilling down into a more detailed or fine-grained view until some target of interest is found. Iterative and interactive information retrieval provides a basis to understand some of the challenges, most notably that of query reformulation, which, without the support of a system, has been known to be a demanding task for users~\cite {belkin2001iterative}. 

Alternatively, this allows users find and pursue a trajectory or a curriculum through a collection to meet a less specific and possibly longer-term goal of learning or familiarizing oneself with a topic. Achieving expertise in a topic does not only involve having access to documents and materials but also acquiring skills that are imparted through conversation and apprenticeship with an expert~\citep{collins2018cognitive,ramberg1998fostering}. 

\subsubsection*{Enriching an interaction with secondary conversational goals}
Among several conversational goals are those of establishing a relationship between counterparts. This is an incidental goal that in human-human interaction is addressed in parallel with informational or narrational activities~\citep{dillard1989primaryandsecondoray,caughlin2010multiplegoals}, which has been studied in human-human communication~\citep{wiemann1977explication}.

Conversational interaction allows users and systems to engage in meta-dialogue to make goals, biases, and backgrounds explicit to themselves and their counterparts. Through conversational interaction, users can assess the reliability and trustworthiness of a system through conversational moves that may not be immediately goal-directed. 

Conversational interaction may add to the appeal of conversation with a system by allowing for ostensibly superficial chit-chat about the topic or theme at hand. This will allow users to gain expertise not only about the topic but also how it can be acted upon.

%% file: section/user.tex
\section{User Aspects}
\label{sec:user-aspects}

In this context, we refer to ``users'' as human agents interested in using an information access system with an \textit{information need}. As we will discuss in~\cref{sec:task}, this need can be diverse, from answering questions, attaining knowledge, planning a trip, or getting recommendations. Users approach systems in the hope that the system will help them meet those needs; to this end, systems offer the user informational, educational, and entertaining materials. 

Some users have preferences, special needs, or other characteristics that will make a conversational system more useful than a non-conversational one; for example, people with visual impairments, 
senior citizens, children, or those seeking coaching. Following principles of Universal Design~\cite{goldsmith2007universal,ostroff2011universal} or ``Design for All'', many or even most such specific preferences will lead to design solutions that improve interaction for every user in the general case. Contextual requirements may benefit from a conversational approach: hands-free operation of a system or situations where a user's visual attention is otherwise occupied are examples where a conversational system would well serve every user.

\subsubsection*{Individual differences}
Users might differ in how much they would like to interact with a conversational system. For example, those with strong domain expertise might have clearer mental models of the domain and can explicate their needs efficiently. A conversational system may impose an annoying or even prohibitive overhead in such cases; expert and experienced users may want to access information through shortcuts---these can be made accessible through conversational systems. Similarly, novices would especially benefit from a conversational interaction as it might help them disambiguate their needs while learning more about the domain through the interaction, with the system being able to ask clarifying questions. It is also important to note that the same user may be a novice in one domain and an expert in another, necessitating a flexible system that can adapt to different levels of expertise across different contexts.

\begin{table}
\centering
\caption{A model of health information seekers, based on \citet{pang2015conceptualising}}
\label{table:PangInfSeekers}
\begin{tabular}{|cc|c|c|} 
\hline
                                  &                    & \multicolumn{2}{c|}{\textbf{Reading Engagement}}              \\
                                  &                    & \multicolumn{1}{c}{\textbf{Low}} & \textbf{High}     \\ 
\hline
\multirow{2}{*}{\textbf{Research Tactics}} & \textbf{Extensive} & All-around Skimmer               & Knowledge Digger  \\ 
\cline{3-4}
                                  & \textbf{Basic}     & Quick Fact Seeker                & Focused Reader    \\
\hline
\end{tabular}
\end{table}

Users also differ in their cognitive needs around information access, e.g., in their tendency to seek information. For example, the health domain~\citet{pang2015conceptualising} categorizes information seekers based on their level of reading engagement and research tactics: a \textit{knowledge digger} would be high on both, whereas a \textit{quick fact checker} would be low on both. In decision-making, a similar distinction is made between users who want to make the best, most optimal decision possible (maximizers) versus those who just want a satisfactory outcome (satisfiers) that satisfies most of their needs. Maximizers usually engage in extensive product comparison and aspire to find the 
best possible option~\citep{Cheek_Schwartz_2016}. In earlier work on a critiquing recommender system, \citet{Frederix2021} found that maximizers experienced more guidance from a (conversational) critiquing system compared to a simple filtering tool, whereas novices did not. Both these examples show that conversational systems might be used, experienced, and thus evaluated quite differently depending on individual differences of the user. Using individual difference measures (such as a maximization scale) allows one to distinguish between users in the evaluation. 

\subsubsection*{Cultural differences}

Cultural differences in how a conversation proceeds, if not accommodated by a conversational system, may annoy and disturb users~\citep{kim1994cross}. For instance, users from cultures that value directness and efficiency might prefer concise, straightforward interactions, whereas those from cultures that emphasize politeness and formality might expect more elaborate and courteous exchanges. This can influence how questions are asked, responses are given, and even the pacing of the conversation. The notion of politeness in itself is a many-faceted behavioral characteristic, and its various aspects are prioritized differently across cultures. 

Moreover, cultural variations extend to preferences in communication styles, such as indirect speech, and non-verbal cues, like pauses or gestures, which a conversational system might need to interpret and adapt to. For example, in high-context cultures, much of the communication relies on implicit understanding, and users might expect the system to read between the lines rather than requiring explicit statements. Conversely, users from low-context cultures might prioritize clarity and explicitness, expecting the system to provide detailed, unambiguous responses.
Language itself is also a significant cultural factor. Variations in dialects, idioms, and slang can pose challenges for a conversational system, which must recognize and appropriately respond to diverse linguistic expressions.

Failure to account for these cultural differences can lead to misunderstandings, frustration, and a decreased sense of trust and satisfaction with the system. Therefore, it is crucial for conversational systems to incorporate culturally adaptive mechanisms, such as language localization, adjustable politeness levels, and the ability to recognize and respect cultural norms and values. During the evaluation, it will also be important also to take such cultural differences into account, as there might be complex interactions between cultural differences and performance and user experience measures.

%% file: section/task.tex
\section{Task}
\label{sec:task}

This section explores the characteristics and tasks suitable for \ac{CONIAC} systems, which are defined by factors such as information need, human involvement, goal orientation, and iterative interaction. In particular, \ac{CONIAC} implies tasks that have certain characteristics, highlighted below
\begin{itemize}
    \item an information need and knowledge gap~\citep{dervin1992sense}
    \item human-in-the-loop (i.e.,\ involving a human user)
    \item goal-oriented
    \item iterative
    \item interactive and mixed initiative~\citep{horvitz1999principles} 
    \item task switching (multiple goals)
\end{itemize}
which we elaborate on in the following.

Traditional information retrieval and recommendation tasks share many of these characteristics, particularly the notion of an information need and a user with a goal that relies on that information.
Some of these tasks are furthered by introducing a conversationally competent system, while others may not be. For example, for a ``factoid'' query with a specific and clear information need, it may not be beneficial for a user to use conversation to satisfy that need. However, a conversational system should still be able to respond to such queries.

In contrast, some tasks strongly benefit from the opportunities afforded by conversation, including an iterative process and agents interacting via turn-taking. Particularly, {\textit{complex}} tasks~\citep{BYSTROM1995191,campbell1988task} that require clarification and refinement to support a user's goal are well-served by an interaction between the system and the user.

\begin{figure}[htb]
    \centering
    \includegraphics[width=0.5\linewidth]{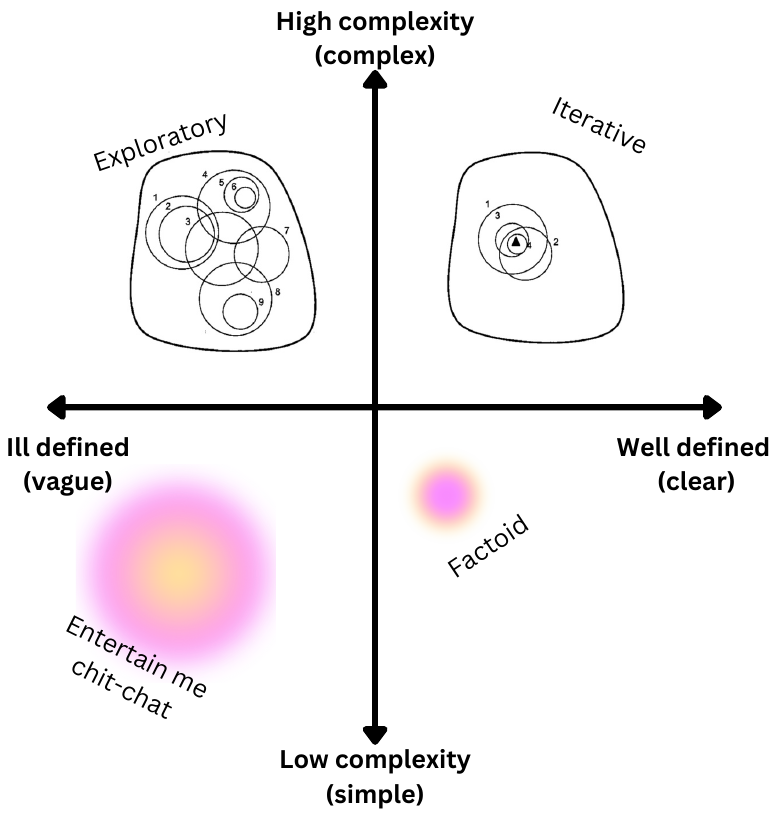}
    \caption{The tasks that are ideally solved by \ac{CONIAC} systems. The model proposes two core dimensions of task complexity and task definiteness. (Iterative vs.\ exploratory search behavior representation adapted from \citet{white2009exploratory})} 
    \label{fig:task_complexity}
\end{figure}

We propose a model for conceptualizing \ac{CONIAC} tasks in~\cref{fig:task_complexity}.

In this model, complex tasks can vary in definiteness, with implications for the nature of an interaction between a system and a user. A complex task that is more well-defined requires primarily an iterative interaction process aimed at refining the expression of the information need to be more focused or precise~\citep{belkin2001iterative}. A more ill-defined information need involves exploration, a process that involves learning and investigation, and examining a broader range of information---often uncovering unanticipated information. This process typically involves refining the information need itself. Here, we draw on the notion of iterative vs.\ exploratory search defined by \citet{white2009exploratory}. This model explicitly reflects the interaction behaviors that are likely to be involved with these two types of search.

A user's information journey in complex tasks is well-documented in the literature~\citep{marchionini2006,kuhlthau1991}, highlighting how user states can evolve throughout the process. For instance, the Information Search Process (ISP) model~\citep{kuhlthau1991} demonstrates that a user's initial perception of a task's complexity might start as simple but can quickly become more complex due to the broad scope of relevant information. This suggests that \ac{CONIAC} systems should be capable of detecting and modeling these evolving user states, optimizing their functions and responses accordingly to enhance the user experience.

Task switching is another common characteristic of human information-seeking behavior, observed in both web search engines and conversational assistants~\citep{SPINK2006264,trippas2024reevaluating}. In complex tasks, users often pursue multiple sub-goals, and the outcome of each sub-goal can influence subsequent actions or significantly impact the final goal. Therefore, it is crucial for users to track multiple tasks effectively. However, traditional information access systems typically leave the burden of managing these tasks to the users.

One example of a complex but well-defined task is the task of buying a smartphone---there are myriad options, with the information spread across many sources and a large and ever-increasing number of variables to consider.

Conversation allows for both personalization and system-human negotiation to define the information need. In short, rather than a one-shot interaction between a user and a system going directly from an articulated information need to an ``answer'', conversation allows for a \textit{process} to arrive at the satisfaction of the user's information need.

In the following, we illustrate typical conversation information access tasks and how they relate to dimensions of complexity and definiteness. We also illustrate how users might dynamically traverse the space during the conversational interaction with the system~\citep{taylor1962,kuhlthau1991}. 

\subsubsection*{Hypothesis generation task}
The generation of hypotheses is a key component of high-level critical thinking in contexts including scientific discovery, problem-solving, and decision-making. Hypothesis generation is a complex information access task. It involves exploring an information space to identify connections between entities, variables, or events that can be tested. It is typically grounded in existing associative or relational patterns. Therefore, it lends itself well to a conversational process. Such an approach helps examine the information space for known connections and then supports a user postulating, prioritizing, and interrogating new ones.

In the context of scientific discovery, hypothesis generation involves investigating scientific literature. This often includes making connections between seemingly unrelated studies~\cite{swanson1987two}. A classic example is the ``ABC'' literature-based discovery model introduced by Swanson~\cite{Smalheiser2017}, in which implicit shared items (``B'' items) are analyzed to form links between two entities (``A'', ``C''). Under this model, a user could be supported by a conversational system to move logically between different parts of the information space.

Evaluation of a hypothesis generation task is challenging as there is no truly ``correct'' answer that the user seeks. Notions of novelty, surprise, feasibility, promisingness, or interestingness are relevant~\cite{smalheiser2012beyond}.

\subsubsection*{Product search and recommendation task}
Suppose a user with limited domain knowledge needs to replace an old washing machine: ``\textit{I need to replace my washing machine and I want one that serves our family of five and is somewhat efficient}''. We initially would classify this as a well-defined need of moderate complexity. Current online tools (e-commerce shops) would allow you to search products with some filtering and construct a set of alternatives, but it will be hard to understand the domain: e.g., what is a sufficient capacity in kg? What amount of kWh makes it efficient? Also, the number of features presented by the tool might bring the realization that there are other important things to consider (rinsing quality, time per cycle) and might overwhelm the user with the number of tradeoffs to make. Hence, a simple search might make the user aware that the need is not as well-defined and the task perhaps more complex than envisioned. However, a regular online shop will not provide the tools to help users understand the domain better. A conversational system, using techniques from critiquing recommenders \cite{Frederix2021} would help the user to understand the domain and tradeoffs better: ``\textit{based on your query, we have several efficient washing machines that have 8kg capacity or more that suit your family needs. Would you rather like a washing machine that rinses the clothes better and but will be less silent}''. Based on this response, the user discovers some other (latent) needs and tradeoffs they were unaware of before, and the dialog will allow them to define them better iteratively. This will allow the user to learn about their actual preferences and the best matching products, and the system can provide a more adequate recommendation.

\subsubsection*{Travel planning task}
This scenario aims to underscore the necessity for a new evaluation framework, criteria, and metrics to examine how complex conversational processes evolve between a user and a \ac{CONIAC} system beyond merely focusing on the outcomes.
A user approached a conversational agent to plan a family vacation. They are four people, including two children and a small dog, with a certain budget level. Initially unsure about destinations, activities, and accommodations, the user relied on the agent’s guidance. The agent recommended Japan---specifically Kyoto and Tokyo---and provided detailed information supported by traveler reviews. When the user expressed concerns about keeping the children engaged, the agent asked probing questions and suggested interactive activities like the Samurai and Ninja Museum, which helped the user feel more confident in the trip’s potential.

As the planning progressed, the user's uncertainty started to diminish. For flights, the agent found a well-reviewed direct option within budget, but detecting the user’s hesitation about long travel times with children, the agent inquired about preferences for layovers or onboard services. This led the user to feel more assured in choosing the direct flight. Regarding accommodations, the user’s initial concerns about the suitability of a traditional ryokan for their children and dog were alleviated after the agent found a more family-friendly alternative. The agent’s ability to adapt recommendations based on evolving user needs gradually transformed the user’s uncertainty into confidence.

By the end of the journey, as the agent reviewed the finalized itinerary, the user had moved from a state of high uncertainty to one of low uncertainty, feeling secure in their travel plans. The user confirmed the itinerary and was satisfied that the agent had addressed all concerns and tailored the trip to their family's specific needs.

\subsubsection*{Health information-seeking task}
Looking for health information is a natural task for~\ac{CONIAC}. Many individuals already use the internet for health advice, such as using symptom checkers to determine if they might have an underlying condition or whether they should consult a healthcare professional~\citep{cross2021search}. 

Research has examined health information-seeking behaviors, for example, considering how users may address an open-ended task such as ``Imagine you are going to a party and will discuss health information with your friends. Gather enough information for an interesting discussion.'' (adapted from \cite{pang2016designing}). This is a vague but complex task, requiring exploration of a large and dynamic information space, accommodation of serendipity to access novel health concepts, and interactions between the user and the system to support the identification of information that is interesting and engaging to the user.

A recent study on a~\ac{CONIAC} system designed to help patients find cancer-related clinical trials indicates that these systems could make health information more accessible for individuals with limited health or computer literacy skills~\citep{bickmore2016improving}. 

While~\ac{CONIAC} has significant potential, several concerns must be addressed when implementing these systems in the health domain. One issue is that these systems may lack the necessary expertise to accurately answer all questions, which could lead to misunderstandings or misinterpretations and, consequently, incorrect responses~\citep{su2019improving}. Although this is a common challenge for all search systems, it could be particularly catastrophic with health information. Additionally, these systems often handle sensitive patient data, which requires robust safeguarding measures. Voice-only~\ac{CONIAC} systems might also face difficulties with speech recognition, especially when users are distressed or in noisy environments~\citep{spina2021SIGIRForum}.

\subsubsection*{Enterprise and personal information management task}

Enterprise information access and personal information access have distinct requirements compared to traditional web search engines. These include challenges such as searching across enterprise (individual) intranets or navigating multiple internal or private data sources~\citep{Hawking2004enterprise}.
Although there has been growing interest in workplace-oriented digital assistants, such as Alexa for Business or the Cortana Skills Kit for Enterprise, their adoption has been limited. However, with the increased adoption of LLM-based systems, new solutions like Microsoft Copilot and Google Gemini are emerging to enhance productivity and streamline information retrieval in enterprise settings~\cite{trippas2024what}.
Microsoft Copilot is designed to assist employees by integrating with Office applications and providing contextual assistance and insights with LLMs. Similarly, Google Gemini aims to leverage artificial intelligence to improve search and information access across Google's ecosystem. Despite these advancements, the use of these tools remains relatively low, suggesting that further development is needed to realize and understand the possible task and the system's potential in the enterprise or personal information environment fully~\cite{trippas2024what, trippas2024reevaluating}.

\subsubsection*{Children's information discovery in the classroom}
Children regularly engage in information discovery tasks that involve exploring online resources to complete classroom assignments. Their default approach is to use web search engines, but they often face challenges: (i)~difficulty in formulating their information needs into succinct keyword queries, (ii)~a preference for using natural language or question-based queries, and (iii)~trouble navigating \acp{SERP} to find relevant resources \cite{azpiazu2017online,landoni2021right,DBLP:conf/sigir/PeraHML25}. A conversational information access system could be highly beneficial in this context. Such a system could ask clarifying questions to help children refine their information needs, provide specific answers when required, and recommend a limited set of resources for further exploration.
However, a one-dimensional evaluation (e.g., the system's 'goodness' based on its ability to deliver the necessary information or user perception—especially, since children's perceptions of their interactions with information systems do not always align with their actions) is insufficient \cite{DBLP:journals/ijmms/BurkeABNKNOPTZ25, DBLP:journals/corr/abs-2405-02050}. As children are the primary stakeholders in this scenario, it is essential to consider factors like the suitability of the information provided and whether the vocabulary matches their literacy levels. This highlights the importance of explicitly considering the task at hand, the stakeholder performing the task, and the various criteria for evaluation before determining the specific metrics to be used as we reflect on the \ac{CAFE} components.

\subsubsection*{Curate a background entertainment task}

Let us suppose that a user starts interacting with a \ac{CONIAC} system as follows: \textit{``The next two and a half hours, I will be working on something that takes up most of my attention, gaze, and hands. I want to have something entertaining in my headphones during that time. I am interested in some general topics: personal aviation, allergy medication, history of Central Asia, and I like yacht rock and string quartets''} (the last part is presumably known to the system). The system can suggest some listening material to the user, excerpts from audiobooks and lectures, recent newscasts with local news, and playlists the user has saved in the library. The user can then respond by ranking the suggestions, helping the system to sort the items that will be played in the session. The system can verify if those preferences are generally true or specific to this session.

\subsubsection*{Longitudinal learning task}

A longitudinal learning task in conversational systems refers to a sustained, adaptive learning process where the system engages with the user over an extended period, facilitating the exploration of various topics or skills. Unlike single-session interactions, longitudinal learning tasks involve multiple interactions spread over days, weeks, or months. These systems are designed to remember past conversations, track the user's progress, adapt to changing interests or knowledge levels, and personalize learning experiences based on ongoing dialogue.
Longitudinal learning tasks in conversational systems can provide personalized and adaptive learning experiences by adjusting to users' evolving interests, mental models, and needs. They enhance knowledge retention through continuous engagement, leveraging past interactions (i.e., memory) for more relevant and context-aware responses.
Evaluating these tasks requires a shift toward more comprehensive, long-term metrics that capture the complexity and evolving nature of sustained learning interactions.

\subsubsection*{Establishing that a system is not the right one for some task}

Let us consider the following scenario. The users asks a system about interesting sights to see in Saarland and bike paths. The user gets some tips, but they do not seem to be geared toward her tastes, so the user elaborates that she wants bike paths for mountain biking and that she will be camping. The system responds but delivers the same content. The user then asks for tips on Limburg, an area she has already vacationed in and know well. The user realizes that the knowledge of this system is superficial while it was initially presented as a useful resource, and she will now turn to another system. (If not conversational, the user would have had to work hard to redo the query for another area).

\subsubsection*{Toward future tasks}

\begin{figure}[htb]
    \centering
    \includegraphics[width=0.5\linewidth]{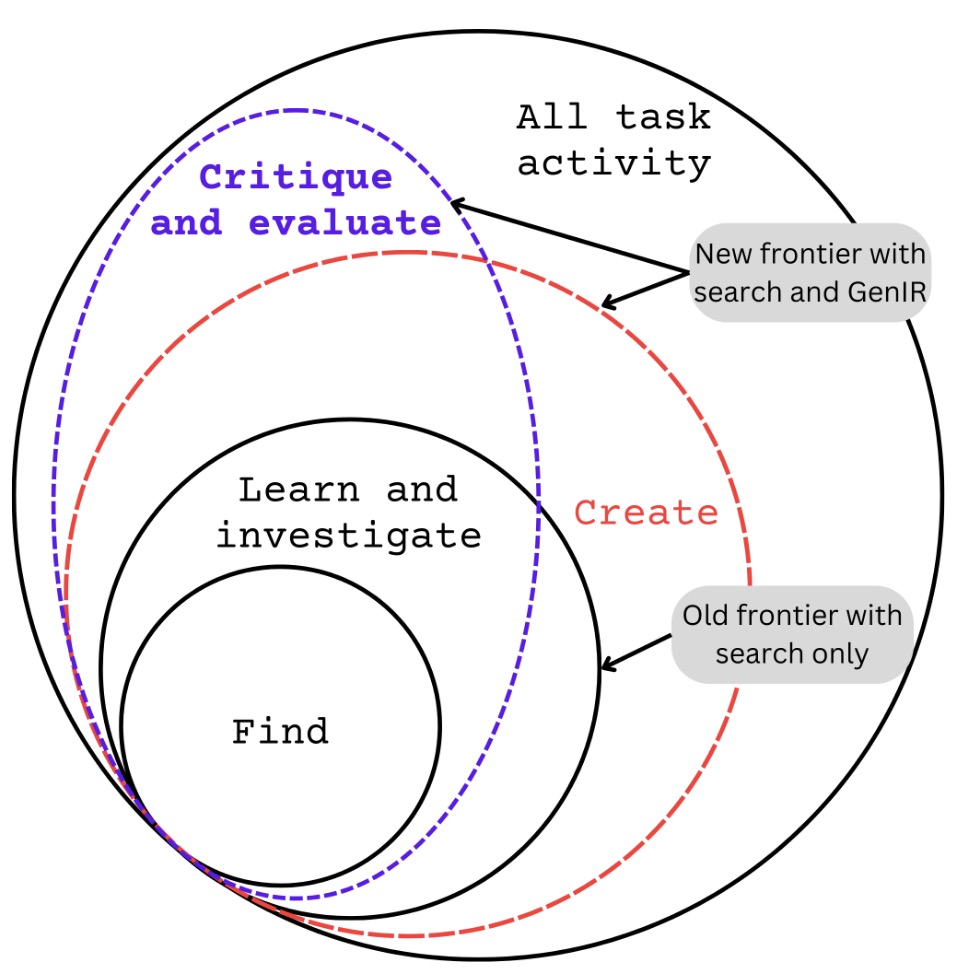}
    \caption{Visualization of possible tasks for information access, demonstrating different information access and activity levels~\citep{white2024tasks, trippas2024adapting}. The diagram highlights the progression from simple information finding to more complex tasks like learning, investigating, critiquing, evaluating, extracting, and creating. The new frontier with Generative Information Retrieval (marked by dashes) indicates that these systems can enable advanced tasks such as critiquing and evaluating, expanding beyond the traditional search frontier also applicable to~\ac{CONIAC} tasks.}
    \label{fig:task_expanding}
\end{figure}

We have yet to exhaustively determine the range of tasks that conversational systems will eventually be capable of performing. 
Recent research indicates that these potential tasks are expanding beyond traditional roles, such as \textit{finding information} and \textit{learning tasks}, to more complex functions like \textit{creating}.
This evolution is illustrated in~\cref{fig:task_expanding}, which shows the shift towards more creative and generative tasks.
We can anticipate future developments in personalized user experiences, enhanced human-AI collaboration, advanced summarization techniques, and multilingual communication. Consequently, the methods for evaluating these evolving tasks will also need to adapt, potentially incorporating new metrics that reflect these systems' increased complexity and capability.

%% file: section/criteria.tex
\clearpage
\section{Criteria}
\label{sec:criteria}

This section discusses the criteria to consider when instantiating an evaluation framework for the conversational search scenario. As mentioned in \cref{sec:goals}, a comprehensive evaluation of the cooperative interaction flow between a \ac{CONIAC} system and its users should involve the goals of different stakeholders. In this sense, the evaluation criteria should also account for the different perspectives on the evaluation, e.g., from a user-oriented, company-focused, or even societal perspective~\citep{jannach2020_mcnamara, bauer2019_multimethod_multistakeholder,
torkamaan2024}. Sometimes, a stakeholder's goals and the corresponding evaluation criteria can overlap and may account for multiple aspects. However, several evaluation criteria can be clearly attributed to the different aspects of the conversational search and recommendation process, being both objective and subjective.

\begin{figure}[htb]
    \centering
    \includegraphics[width=0.95\linewidth]{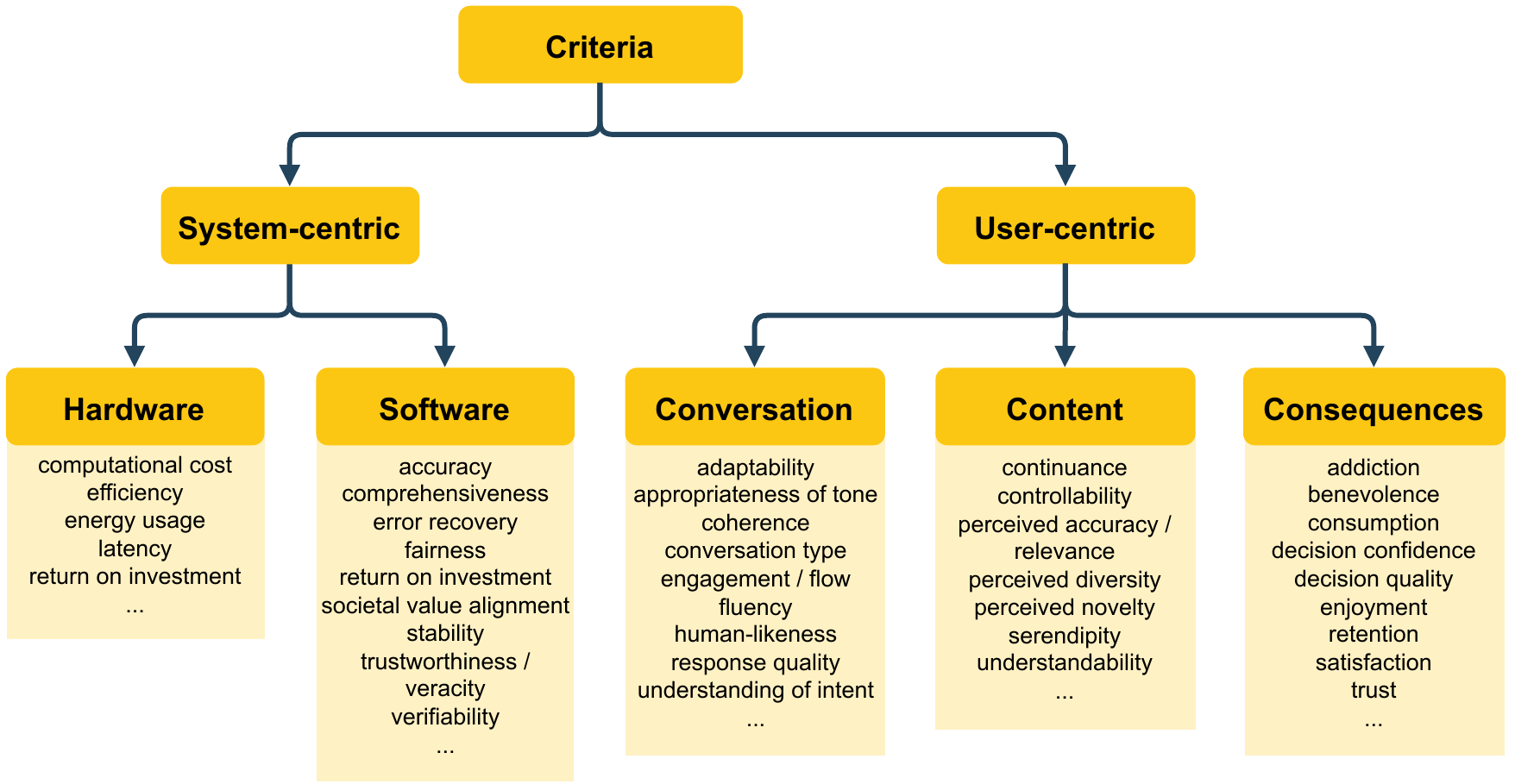}
    \caption{Taxonomy of evaluation criteria.}
    \label{fig:criteria}
\end{figure}

\cref{fig:criteria} illustrates a taxonomy of criteria that can be used to evaluate a conversational system. The first partitioning of the criteria concerns the subject of the evaluation, and it allows us to identify two major classes of evaluation criteria: \textit{System-centric} and \textit{User-centric} evaluation criteria. Note that the literature about criteria is rich and diversified, lacking standardized classifications for them across different research areas. It is beyond the scope of this manifesto to propose such a commonly agreed classification. Therefore, by no means do we claim the listed criteria in the taxonomy to be exhaustive, but rather, we aim to categorize the different aspects of the cooperative conversational process, providing examples of suitable evaluation criteria.

\subsection*{System-centric criteria}

With system-centric evaluation criteria, we refer to those aspects typically---but not exclusively---characterized by a high level of objectivity. These criteria include those that drive the development of the system (i.e., they can be derived from the stakeholders' constraints or induced by the goals of the system) and those that can be evaluated in offline scenarios using the system in isolation, although they may require some preexisting user input, such as existing user ratings or expert labels.
We can further partition system-centric criteria into \textit{hardware} and \textit{software} criteria.

\subsubsection*{Hardware-centric criteria}
Hardware-centric evaluation criteria include, among others, the system's energy usage, computational costs, efficiency, and latency. Modern state-of-the-art \ac{CONIAC} systems mostly rely on \acp{LLM} with dedicated hardware and energy requirements. As earlier work by \citet{DBLP:conf/acl/StrubellGM19} pointed out, the energy usage of \acp{LLM} has a considerable environmental impact, and these implications should be kept in mind by platform operators, system developers, and also by the users. Closely related to energy usage, computational costs are a more general criterion that is of special interest to the platform operators and service providers, who eventually have to cover the resulting financial costs. Regarding the system's response time, latency and efficiency are important criteria for system operators and users alike. Long response times may dissatisfy the users, making the service less attractive, which can potentially lead to user abandonment or quitting the use of the service entirely.

\subsubsection*{Software-centric criteria}
Software-centric criteria include those criteria that are not particularly bound to the hardware. Among others, the system's accuracy, comprehensiveness, or error recovery are objective criteria that can be quantified to better understand the \ac{CONIAC} system's impact on the conversation when evaluating the conversational process. For instance, \textit{accuracy} reflects how well the system provides correct responses (e.g., accurate recommendations) to the user's request. \textit{Comprehensiveness} accounts for how thoroughly the request is answered and if it requires follow-up interactions like rephrasing or making more utterances in general. 
\textit{Error recovery} refers to the quality of recognition and recovery from misunderstandings on both sides of the conversation.

Other criteria---of particular interest to the society as a whole---include fairness, replicability, and the alignment to societal values. An evaluation of fairness criteria allows for a better understanding of potential biases in the data or the algorithms. Similarly, evaluating the replicability of system-based components allows audits of the reliability and robustness of the \ac{CONIAC} system, which may inherently impact other criteria like the system's trustworthiness. In this regard, the development of the system can also be driven by societal values, which, in turn, requires an evaluation of how well the \ac{CONIAC} system is able to grasp and convey these values to the users in the end. 
In this context, \textit{veracity} plays an important role, which accounts for the system's ability to provide responses that also reflect reality. This is especially important, considering the consequences for decision-making contexts. Furthermore, recent work by \citet{DBLP:conf/sigir/JokoCRV0H24} suggests that \ac{LLM}-based conversational agents are not diverse, so \textit{diversity} can be an important criterion for future evaluations of \ac{CONIAC} systems.

We refer the reader to \citet{DBLP:journals/corr/abs-2305-08290}, who also provides an elaborated taxonomy of criteria for evaluating textual \ac{CONIAC} systems. Many of the criteria mentioned by \citet{DBLP:journals/corr/abs-2305-08290} can be aligned with the different aspects we envision in our taxonomy.

\subsection*{User-centric criteria}
User-centric criteria represent the second broader category of aspects that should be evaluated when it comes to interactive systems~\citet{DBLP:journals/ftir/Kelly09,knijnenburg2012} and especially \ac{CONIAC} systems~\cite{jin2024}. 
Users' subjective reflections on their interaction with the system can be characterized as their perceptions of certain qualities of the system itself and the self-relevant consequences of their use of the system. For \ac{CONIAC} systems, we divide users' perceptions of system qualities into perceptions of the \textbf{conversation} itself (e.g., engagement, response quality), and perceptions of the \textbf{content} that is recommended/retrieved by the system (e.g., perceived accuracy, serendipity).

The \textbf{consequences} of users' interaction with the system have both a behavioral and subjective component to them---the subjective component concerns users' satisfaction with the outcome of system use (e.g., decision confidence), their satisfaction with the process of using the system (e.g., fun), and their trust in the system itself.
The behavioral component concerns both users' interactions within the system (e.g., consumption, retention), as well as actions that a user takes in the real world (e.g., recommending the system to friends), and the consequences thereof (e.g., decision quality).

\subsection*{An emphasis on temporal evaluation}

It is important to note that the \textbf{temporal} dimension of the interaction is particularly important in \ac{CONIAC} systems---this relates to both the \emph{flow} of the conversation between the user and the system, the \emph{continuance} of the cooperative interaction between the user and the system towards fulfilling the information access goal, and the development of the user's \emph{trust} in the system~\cite{kahr2024-2,DBLP:journals/sigir/AzzopardiCKMTRAACREHHKK24}. This means that evaluating \ac{CONIAC} systems requires us to evaluate the system at various temporal scopes: evaluation can address a single turn (i.e., the traditional evaluation metrics), a conversation (e.g., flow, continuance), or the system usage across multiple conversations (e.g., trust). Measuring the interaction at all of these levels means applying methodologies and metrics that extend beyond a single interaction and that do not just cover a \ac{CONIAC} systems' stationary properties but also its dynamic properties (which can be measured either continuously or at a certain discrete temporal rate).

%% file: section/methodology.tex
\section{Methodology}
\label{sec:methodology}

The advantage of \ac{CONIAC} systems is their support for complex, rich, cooperative interactions that go beyond the back-and-forth interactions with traditional \ac{IR} and \ac{RS} systems. This richness of conversations must hence be reflected in the evaluation of \ac{CONIAC} systems, even beyond extensions for generative systems like that of \citet{gienapp:2024a}. Traditional interactive \ac{IR} and \ac{RS} evaluations span from system-focused evaluations to user-focused evaluation methodologies \cite{DBLP:journals/ftir/Kelly09,zangerle2022_fevr}---in the context of \ac{CONIAC} systems, we advocate for an even stronger emphasis on user-focused evaluation to better capture the qualities of these complex, cooperative interactions. Additionally, the dimension of time is critical for \ac{CONIAC} systems, which is not reflected in most \ac{IR} and \ac{RS} evaluation methodologies used so far.

Therefore, we categorize evaluation methodologies---both quantitative and qualitative---along two dimensions: (1)~according to the focus of a study (a standard dimension in \ac{IR}), ranging from system-focused methodologies like offline simulations to user-focused methodologies like qualitative interviews; and (2)~according to the employed time model (a dimension from system behavior theory that is especially suitable for \ac{CONIAC} studies), ranging from stationary methodologies like single-interaction experiments to methodologies like longitudinal user studies that allow for continuous measurements of, for example, satisfaction. While evaluation criteria are defined as agnostic of time models, for the actual evaluation, one has to decide which time model to use, which limits the available evaluation methodologies and affects the details of their implementation. Given the criteria of interest, one has to choose an evaluation methodology that covers these criteria. In this respect, we give a brief overview of existing methodologies, highlighting the clear divide between user-centric and system-centric approaches.

\subsection*{Methodologies for \ac{CONIAC} systems}

\begin{figure}[tb]
\centering
\includegraphics[width=\linewidth]{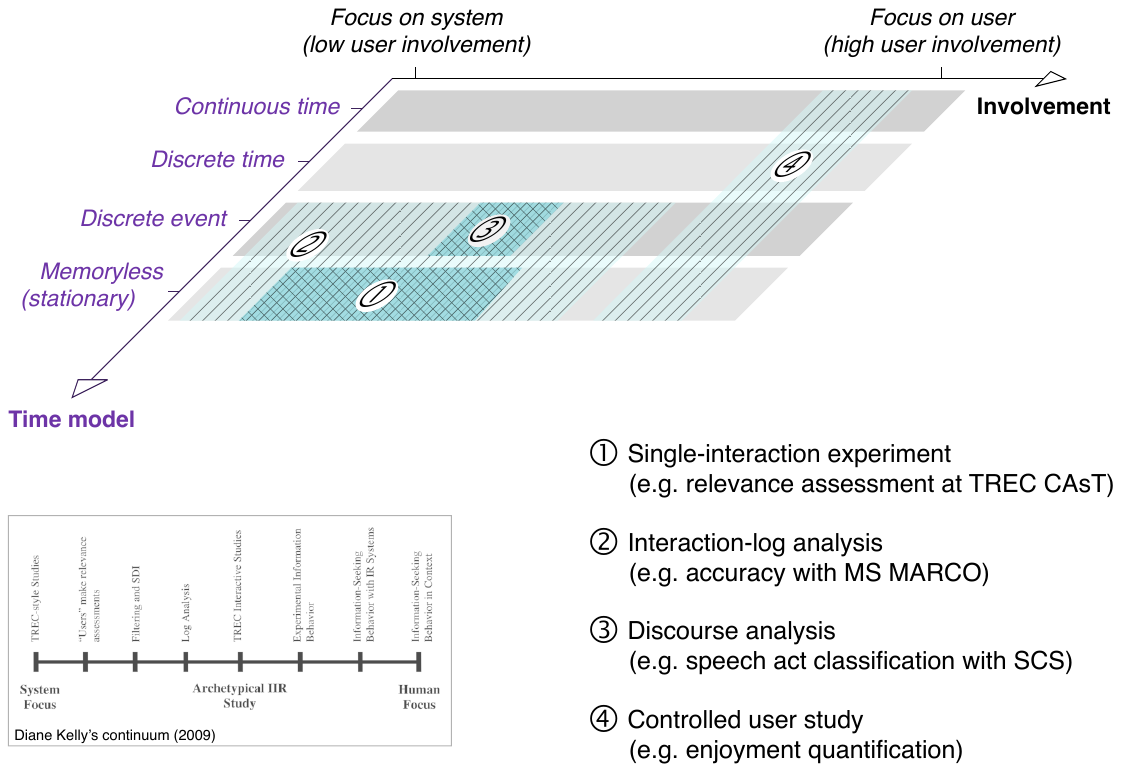}
\caption{In conversational settings, time makes the difference. See above the methodologies spectrum, ``Involvement'' (the x-axis) as introduced by Kelly~\cite{DBLP:journals/ftir/Kelly09} (shown lower left), extended by an orthogonal ``Time model'' axis. The figure also gives four examples of methodologies and how they cover different parts of the user involvement and time model space.}
\label{fig:cia-spectrum}
\end{figure}

Evaluation methodologies for information systems can be placed along a continuum \cite{DBLP:journals/ftir/Kelly09} from focusing on the system (without integrating actual users, like offline simulation) to focusing on the users (without integrating specific systems, like behavioral diary studies~\cite{teevan:2004}). Methodologies focusing on the interaction between specific users and specific systems are placed in the center. The space depicted in \cref{fig:cia-spectrum} has this continuum as its horizontal axis.

Conversations are complex, dynamic interactions that evolve continuously as they progress toward the user's goal. 
This temporal aspect corresponds to two different concepts: 
The \textit{scope} (how much we measure, from single turns to multi-conversation interactions) and the \textit{model of time} (unit of analysis, from single discrete to continuous).
We expand the one-dimensional continuum of \citet{DBLP:journals/ftir/Kelly09} with the time model that the methodology implements, taken from the theory of system behavior~\cite{wymore1976systems} (cf.\ \cref{fig:cia-spectrum}).
Therefore, evaluation methodologies must be selected or adapted to fit the concept of time the research question demands.
The four different time models, each implemented by several methodologies, are the following:

\subsubsection*{Memoryless (stationary)}
A single data point is measured per conversation (named ``memoryless'' in system behavior theory). For some research questions, especially when comparing systems, it is sufficient to measure either only once for a conversation---typically at the end---or to take only an aggregated measurement like an average. A typical example of system-focused evaluations with this time model are single-interaction experiments, which include the TREC~CAsT~\cite{dalton2020treccast2019conversational,DaltonEtAl2020,dalton:2021,owoicho:2022} and TREC~iKAT~\cite{aliannejadi:2023} competitions in which systems are asked to respond to a user utterance with conversation history, or the Netflix prize in which algorithms are asked to increase recommendation performance across a dataset~\cite{bennett2007}. A typical example of user-focused evaluations with this time model are controlled user studies with a post-conversation survey~\cite{knijnenburg2015}.

\subsubsection*{Discrete event}
The occurrence of discrete events is measured.
Typically employed events are the utterances in conversations with chat-based agents and speaker changes for voice-based agents. Other suitable resolutions are single words uttered, topic switches, when the user has completed some sub-tasks and asks the agent for new instructions (e.g., when the agent assists in cooking), or when the recommended song is finished. A typical example of evaluation methodology is interaction-log analysis, where the events are reconstructed from the logs. Examples of available search interaction-logs include MS~MARCO~\cite{nguyen:2016} Conversational Search\footnote{An artificial log generated based on real search engine usage\\\url{https://microsoft.github.io/MSMARCO-Conversational-Search/}} and CIRQL~\cite{trippas2024reevaluating}. 
Detailed behavioral logs may be created by employing advanced user tracking mechanisms~\cite{willemsen2011}. If no logs of real interactions are available, offline simulation can be employed to generate interactions, for example, using software like SimIIR~\cite{zerhoudi:2022}, GenIRSim~\cite{kiesel:2024c}, and UserSimCRS~\cite{DBLP:conf/wsdm/AfzaliDB023}.

One evaluation methodology that follows a discrete event time model that is especially relevant to \ac{CONIAC} is discourse analysis, which entails parsing the conversation according to theories from the discourse and dialog literature as an intermediary step for the evaluation. The parsed representation of the conversation follows, for example, rhetorical structure theory~\cite{mann:1988}, models of argumentation~\cite{stede:2018}, or the conversational roles model~\cite{sitter:1992}---the last was specifically developed for \ac{CONIAC} and builds on top of the theory of speech acts~\cite{searle:1969}. The parsed representation can then be employed as a model of the conversation~\cite{KieselEtAl2021}, for example, to detect anomalous conversations~\cite{vakulenko:2019}. Available datasets include SCS~\cite{trippas2020towards} and MANtIS~\cite{penha:2019}.

\subsubsection*{Discrete time}
Measurements are taken at discrete and regular points of time according to a selected sampling rate, ranging from a fraction of seconds to days. As conversations are typically analyzed in terms of the events described above, this time model is not as frequent in \ac{CONIAC} experiments. However, especially for wearable or ubiquitous conversational agents like ones integrated into smart watches or glasses, experience sampling~\cite{pejovic2016}---which uses the discrete time model by definition---could be an especially suitable methodology to gain insights into natural usage behavior. Discrete-time evaluations may also be used in longitudinal studies to track the evolution of users' interactions with the system. Finally, in descriptive research, diary studies can be used to obtain user input at set time intervals, e.g., on a daily basis~\cite{carter2005}.

\subsubsection*{Continuous time}
Measurements are taken continuously. Though measurements taken with computers might not be genuinely continuous, methodologies that use this time model could use a very high sampling rate to effectively resemble continuous measurements and thus allow for calculating measurement derivatives. Physiological measurements are often taken continuously, for example, saccades and dwell times in eye tracking studies~\cite{beckers:2010}. Continuous measurements do not need to be without interruption. When evaluating several interaction sessions with \ac{CONIAC} systems, the measurement can pause. 
\subsection*{Existing methodologies}
In the following, we briefly define well-known methodologies that can be used to evaluate \ac{CONIAC} systems, some of which have been mentioned above.

\begin{description}
\item[Single-interaction experiment] The class of offline methodologies that examine a single user interaction with a conversational system. This technique is among the easiest methodologies for rapid assessment.
\item[Offline simulation (incl. A/B)] Offline simulation involves testing a system or model using pre-recorded data instead of real-time data~\cite{DBLP:conf/sigir/HsuMMPPLLPYSB24}. Offline A/B testing compares two versions of a system by running simulations on historical data to see which performs better. These methods help in evaluating changes without affecting live users.
\item[User simulation (incl.\ LLM-based)] User simulation refers to methods where user behaviors are simulated by software. Especially, \ac{LLM}-based approaches allow for simulating users' interaction behaviors for various tasks such as conversational search and recommendation~\cite{sekulic2024,Zhang2024simulate}. 
\item[Online A/B testing] Online A/B testing involves comparing two versions of a system to see which performs better~\cite{kohavi2009}. Users are randomly shown either version A or version B, and their interactions are tracked to measure effectiveness and/or efficiency.
\item[Interaction-log analysis] Interaction-log analysis involves studying past user interactions to understand how people use a system. By examining these logs, patterns, and trends can be identified in user behavior. This information helps improve the conversational system to meet user preferences better.
\item[Expert evaluation] Expert (or editorial) evaluation involves having specialists or experienced editors assess components of conversations. The experts provide feedback based on their knowledge of the field. 
\item[Discourse analysis] Discourse analysis is a research field for the offline study of transcripts from written or spoken conversations. Discourse analysis can be done on various levels of abstraction, ranging from the very fine-grained \textit{conversation analysis} family of methods~\cite{sidnell2012handbook} to methods based on, e.g., rhetorical structures, social processes, or systemic linguistics.
In general, discourse analysis examines how linguistic and, in some cases, para-linguistic mechanisms and elements are used to achieve communicative goals in recorded interactions and how they make it possible for participants to make sense of a conversation~\cite{grosz1986attention}. 
\item[Data donation] Data donation methodologies involve different ways people can share their data for research or analysis~\cite{DBLP:conf/chi/ZannettouNAGGRR24}. This can include directly providing data, allowing access to existing data, or contributing anonymized data collected by devices. These methods help researchers gather valuable information while respecting privacy and consent. They also allow the gathering of information that sits across multiple systems or authorities.
\item[Wizard of Oz / observational study] The Wizard of Oz methodology involves creating a prototype where users interact with a system they believe to be automated but can be operated by a human behind the screen~\cite{dahlback1993,knijnenburg2016}. This enables researchers to test and refine the system's design and functionality before fully developing the technology. Alternatively, for conversational experiments, two humans could observe their interactions as a first step, referred to as observational studies~\cite{trippas2020towards}. It is a cost-effective, rapid deployment way to gather user feedback and improve the system based on real interactions.
\item[Controlled user study (incl.\ longitudinal)] Controlled user studies involve observing and testing users in a structured environment to evaluate how they interact with a system~\cite{knijnenburg2015}. Longitudinal studies extend this by following the same users (commonly referred to as panels) over a longer period to see how their interactions and experiences change over time. Commonly explicit feedback is gathered after the study. These methods help researchers understand both immediate and long-term effects.
\item[User survey (quantitative)] User survey methodologies involve collecting feedback from people through questionnaires. These surveys ask users to outline their experiences, preferences, and opinions to gather valuable insights.
\item[User interview (qualitative)] Qualitative user interview methodologies involve conducting in-depth conversations with users to understand their experiences, thoughts, and feelings. These interviews typically follow a methodology (e.g., grounded theory~\cite{charmaz2015grounded}). This approach helps researchers gain an understanding of motivations.
\item[Diary study~/ experience sampling] Diary study and experience sampling methodologies involve participants reporting their thoughts, feelings, and activities. In the former, participants write diary entries. In the latter, participants are prompted throughout the day. This approach helps researchers understand people's experiences and behaviors in their natural environments.
\end{description}

\begin{table}[tb]
     \centering
     \resizebox{\textwidth}{!}{
     \begin{tabular}{@{}lccccc@{}}
     \toprule
        \textbf{Methodology} & \multicolumn{5}{c}{\textbf{Criteria}} \\
     \cmidrule(l){2-6}
        & \multicolumn{2}{c}{\textbf{System-centric}} & \multicolumn{3}{c}{\textbf{User-centric}} \\
     \cmidrule(lr){2-3}
     \cmidrule(l){4-6}
         & \textbf{Hardware} & \textbf{Software} & \textbf{Conversation} & \textbf{Content} & \textbf{Consequences} \\
     \midrule
        single-interaction experiment                         & \checkmark &      *      &            &            &            \\
        offline simulation (incl. A/B)              & \checkmark & \checkmark &            &            &            \\
        user simulation (incl. LLM-based)  & \checkmark & \checkmark &            &            &            \\
        online A/B testing                          &     *       & \checkmark &            &            &      *     \\
        interaction-log analysis            & \checkmark &       *     &     *      &            &      *     \\
        expert evaluation                           &            &            &     *      &      *     &            \\
        discourse analysis                          &            &            & \checkmark &     *       & \checkmark \\
        data donation                               &            &            &            &            &      *     \\
        wizard of oz / observational study                          &            &            & \checkmark &     *       &      *     \\
        controlled user study (incl. longitudinal)  &     *       &   *         & \checkmark & \checkmark & \checkmark \\
        user survey (quantitative)                  &            &            & \checkmark & \checkmark &      *     \\
        user interview (qualitative)                &            &            & \checkmark & \checkmark &      *     \\
        diary study / experience sampling           &            &            & \checkmark & \checkmark & \checkmark \\
        \\
        \bottomrule
     \end{tabular}}
     \caption{Examples of methods suitable for evaluating criteria categories (\checkmark---all criteria of this category, *---only some)}
     \label{tab:criteria-methods}
\end{table}

For planning an evaluation, one usually starts from the goals of the evaluation (see~\cref{sec:goals}) and then defines the criteria of interest (see~\cref{sec:criteria}). In the next step, one needs to choose an evaluation methodology covering these criteria. \cref{tab:criteria-methods} shows the methodologies described above and lists the criteria categories of~\cref{sec:criteria} for which they are most suitable. As the table shows, there is a mostly clear divide between system- and user-oriented methodologies \cite{DBLP:journals/ftir/Kelly09}.

%% file: section/measure.tex
\section{Measures}
\label{sec:measure}

This section presents the commonly used measures for assessing system-centric hardware and software criteria, and the measures for evaluating user-centric criteria (including users' subjective reflection on their interactions, their perceptions, and behavioral measures). Depending on the goals and criteria a specific system aims to fulfill, one can adopt the suitable evaluation methodology (see examples in \cref{tab:criteria-methods}) to assess the corresponding measures.

\subsection*{Measures for system-centric criteria}

There are tutorials and reviews of measures in \ac{IR}~\cite{Sanderson2010}, \ac{RS}~\cite{Gunawardana2022}, and software~\cite{FentonBieman2014}. As with all measures, it is important to ensure they are used with care~\cite{Fuhr2017}.

Measures are commonly used to quantify criteria. For hardware, common measures include computational cost, potentially specified in currency units; efficiency, potentially specified in terms of computational needs per data unit processed; energy usage, potentially specified in kilowatt hours per computational task; and latency, potentially specified in seconds.

Software measures include, but are not limited, to accuracy, potentially specified in \ac{nDCG} or \ac{AUC}; comprehensiveness, potentially specified in terms of recall; error recovery, potentially specified in terms of mean time to repair; fairness, potentially specified in terms of the divergence from expected distribution; replicability, potentially specified in terms of the ease with which others can reproduce an existing algorithm/system; 
stability, potentially specified in terms of mean time between failure;
veracity, potentially specified by data error rate;
and verifiability, potentially specified by the percentage of fact indication covered by evidentiary sources. 

\subsection*{Measures for user-centric criteria}

Measuring the effects of a \ac{CONIAC} system on its users requires either observing the user's interaction with the system or probing the user for their reflection on their interaction with the system. 

Users' subjective reflection on their interaction with the system can be measured using multi-item measurement scales, many of which can be captured by existing user-centric evaluation frameworks~\citep{jin2024,knijnenburg2012} or established scales for specific aspects (e.g., for various dimensions of trust~\citep{mcknight2002}. It is important to adapt these scales to the context of the system (e.g., a scale for goal achievement should broadly refer to the goal of the specific system under evaluation, if known) or to develop new scales where needed---the interested reader can refer to~\citet{devellis2011} for a primer on how to develop robust subjective measurement scales.

Behavioral measures that comprise users' interactions with the system can be measured using conversation logs and compiled into measures that have been defined extensively in the area of A/B testing~\citep{kohavi2009}. Objective outcomes that occur outside the system have been covered in the domain of behavioral psychology~\citep{winkler1973}. 

It is important to note that users' perceptions and behaviors (both within and outside the system) are subject to 
be affected by users' personal characteristics and contextual variables~\citep{Cai2022trust}.
It is also important to measure users' perceptions and behaviors at the appropriate temporal scale and/or frequency. For example, users' decision-making can be traced temporally through expansive interaction logging~\cite{schulte2017}, and the development of trust within an interaction session can be measured by repeatedly probing users with trust-related questions or by repeatedly observing trust-related behaviors~\cite{kahr2024,kahr2024-2}.

%% file: section/research.tex
\section{Research Directions}
\label{sec:research}

In this section, we highlight some possible questions that remain open and may constitute future research directions, requiring further and deeper investigation.

\subparagraph{Stakeholder Goals}

 \begin{itemize}
    \item What do users really expect from \ac{CONIAC} systems?
    \item How can \ac{CONIAC} systems accurately deduce stakeholders' goals from interactions with users?
    \item How can we help users select the best \ac{CONIAC} for a given task?
\end{itemize}

\subparagraph{User Aspects}

\begin{itemize}
    \item Which types of users can benefit substantially from \ac{CONIAC} systems and how?
    \item What are good measures of individual and cultural differences for users of \ac{CONIAC} systems?
    \item What individual and cultural differences are important moderators of the success of (specific types of) \ac{CONIAC} systems?
    \item How do we ensure statistical validity and power when experimenting with different types of users?
\end{itemize}

\subparagraph{Task}

\begin{itemize}
    \item What common characteristics do complex tasks have that can benefit significantly from \ac{CONIAC} systems?
    \item How can \ac{CONIAC} systems become closely integrated into task environments, and how can their contribution to task performance be evaluated in these settings?
    \item What temporal patterns can be observed in the level of perceived task complexity during interactions with a \ac{CONIAC} system?
    \item How can a user and a \ac{CONIAC} system efficiently agree that a given task is beyond the system's capabilities?
    \item What array of criteria and methods can capture, account for, and reflect the dynamic, evolving, and possibly multi-modal nature of conversations?
\end{itemize}

\subparagraph{Criteria}

 \begin{itemize}
    \item What criterion makes a difference in users' evaluation of \ac{CONIAC} systems?
    \item Can we foresee a future when we will be able to assess user-centric criteria (e.g., during system development) using \acp{LLM} in a simulation setting? 
    \item How can we structurally report on multi-criterion evaluations that include both system-centric and user-centric criteria?
\end{itemize}

\subparagraph{Methodology}

 \begin{itemize}
    \item What methods can be developed to evaluate \ac{CONIAC} systems that prevent test data leakage while ensuring robust and reliable evaluation metrics?
    \item How can we accurately attribute the performance of a conversational model to its individual components? 	
    \item What approaches can be used to evaluate the accuracy and relevance of responses from a \ac{CONIAC} system, particularly when these responses are derived from the aggregation of multiple documents and the system’s internal knowledge?
    \item Can we reuse interactions and corresponding labels between a system and a user to evaluate a different system, or would this bias our observations on the second system?
    \item How do we ensure reproducibility of experiments when they involve users, interaction, possibly simulation, and inherently stochastic systems (e.g., \acp{LLM})?
\end{itemize}    

\subparagraph{Measures}

\begin{itemize}
    \item Besides user studies, what offline evaluation measures can we develop to assess continuous criteria such as conversation flow and continuance?
    \item What is the most effective and statistically sound approach to aggregate measures across different interactions? Is it just averaging, or should we develop protocols more aligned with the structure of a conversation (e.g., non-independent interactions, graph structure)?
    \item Which measures are suited to evaluate the dynamic evolution of a conversation over time, including changes in mental models and user-centric evaluation criteria?
\end{itemize}

%% file: section/conclusions.tex
\section{Conclusions}
\label{sec:conclusions}

In this manifesto, we made a case for the need for more advanced conversational systems (\cref{sec:introduction}), which go beyond what current technology---despite being already very performing---is able to do. In this respect, we elaborated an abstract and essential \emph{World Model} (\cref{sec:coniac}) for the area we called \acf{CONIAC}. In this model, we defined how a conversation goes on as a sequence of events in the \ac{CONIAC} process layer and introduced the main features of \ac{CONIAC} systems in terms of: (i) blending different technological domains, such as \ac{IR} and \ac{RS}; (ii) embedding both internal knowledge about the user and external knowledge about the world or context in which the user and the system operate; (iii) tracking the state of the conversation both as flows of events and as dynamically updated beliefs in the user information state.

The above vision of the \ac{CONIAC} world led us to elaborate the \acf{CAFE} (\cref{sec:cafe}). This stems from the observation that, to align with the flows of the events in the \ac{CONIAC} process layer and with the evolution of the state of a \ac{CONIAC} system, evaluation should turn into a dynamic series of probes rather than being a single static assessment, as it is mostly today.

Then, in the following sections, we detailed the main areas of the \ac{CAFE} framework, namely, Stakeholder Goals (\cref{sec:goals}), User Aspects (\cref{sec:user-aspects}), Task (\cref{sec:task}), Criteria (\cref{sec:criteria}), Methodology (\cref{sec:methodology}), and Measures (\cref{sec:measure}). We provided suggestions and examples of possible instantiations for each of these areas without pretending to be exhaustive.

Finally, we presented open research questions that have emerged during the discussions on the different parts of \ac{CAFE} that may suggest future research directions.

This manifesto should not only be regarded as a useful account of an important research challenge. We hope that it will also produce valuable fall-outs, such as bringing these issues into the research agenda of the involved communities, providing a common ground to coherently develop cross-sectorial research, helping funding agencies envision appropriate funding instruments for addressing these challenges, and motivating researchers to overcome today’s limitations.

%% file: section/acknowledgements.tex
\section*{Acknowledgements}

We thank Schlo{\ss} Dagstuhl for hosting us.

\noindent Icons are taken from the Noun Project\footnote{\url{https://thenounproject.com/}}.

\noindent This work was partially supported by CAMEO\footnote{\url{https://cameo.dei.unipd.it/}}, PRIN 2022 n. 2022ZLL7MW, and by the Excellence in Digital Sciences and Interdisciplinary Technologies (EXDIGIT) project, funded by Land Salzburg under grant number 20204-WISS/263/6-6022.